\documentclass{article}

\usepackage[final, nonatbib]{neurips_2023}

\usepackage[utf8]{inputenc} 
\usepackage[T1]{fontenc}    
\usepackage{hyperref}       
\hypersetup{breaklinks=true,colorlinks}
\usepackage{url}            
\usepackage{booktabs}       
\usepackage{amsfonts}       
\usepackage{nicefrac}       
\usepackage{microtype}      
\usepackage{xcolor}         

\usepackage{graphicx}
\usepackage{amsmath}
\usepackage{amssymb}
\usepackage{booktabs}
\usepackage{enumitem}
\usepackage{makecell}
\usepackage{booktabs}
\usepackage{float}
\usepackage{algorithm}
\usepackage{algpseudocode}
\usepackage{xspace}
\usepackage{colortbl}
\usepackage{times}
\usepackage{epsfig}
\usepackage{graphicx}
\usepackage{amsmath}
\usepackage{amssymb}
\usepackage{subcaption}
\usepackage{pifont}
\usepackage{amssymb}
\usepackage{amsthm}
\usepackage{thmtools}
\usepackage{thm-restate}
\usepackage{dsfont}
\usepackage{footmisc}
\usepackage{caption}

\definecolor{Gray}{gray}{0.85}
\newcolumntype{a}{>{\columncolor{Gray}}c}

\usepackage{amsmath}
\usepackage{fancyhdr}

\usepackage[capitalize]{cleveref}
\crefname{section}{Sec.}{Secs.}
\Crefname{section}{Section}{Sections}
\Crefname{table}{Table}{Tables}
\crefname{table}{Tab.}{Tabs.}

\newcommand{\xmark}{\ding{55}}%
\newcommand{\RR}{{\mathbb{R}}}
\newcommand{\PP}{{\mathbb{P}}}

\newcommand{\EE}{{\mathbb{E}}}

\newcommand{\calD}{{\mathcal{D}}}
\newcommand{\Wb}{{\mathbf{W}}}

\newcommand{\ca}{\operatorname{cross-attn}}
\newcommand{\soft}{{\operatorname{softmax}}}
\newcommand{\mysign}{{\mbox{sign}}}
\newcommand{\xb}{{\mathbf{x}}}

\newcommand{\inca}{\operatorname{InCA}}
\newcommand{\openinca}{\operatorname{Open-InCA}}
\newcommand{\parens}[1]{\left(#1\right)}
\newtheorem*{definition*}{Definition}
\newtheorem{definition}{Definition}

\declaretheorem[name=Theorem, numberwithin=section]{theo}

\declaretheorem[name=Observation, sibling=theo]{obs}

\definecolor{yd}{rgb}{0.0, 0.5, 0.0}
\definecolor{codecolor}{rgb}{0.5, 0.55, 0.9}
\def\name{InCA\xspace}

\newcommand{\ie}{\textit{i.e.}}
\newcommand{\eg}{\textit{e.g.}}

\newcommand{\norm}[1]{\left\lVert#1\right\rVert}

\usepackage{multicol} 
\usepackage{titlesec}
\titlespacing{\paragraph}{0pt}{0.5\baselineskip}{0.5\baselineskip}

\title{Your representations are in the network: composable and parallel adaptation for large scale models}

\author{Yonatan Dukler \And Alessandro Achille \And Hao Yang \And Benjamin Bowman\thanks{Work conducted while interning at AWS AI Labs.}~ \And Varsha Vivek \And Luca Zancato \And
Avinash Ravichandran \And Charless Fowlkes \And  Ashwin Swaminathan \And Stefano Soatto \AND  AWS AI Labs \\
\tt\small \{dukler, aachille, haoyng, bowmaben, varshviv, zancato, \\
\tt\small ravinash, fowlkec, swashwin, soattos\}@amazon.com}

\begin{document}

\maketitle

\begin{abstract}
We present a framework for transfer learning that efficiently adapts a large base-model by learning lightweight cross-attention modules attached to its intermediate activations. 
We name our approach InCA (Introspective-Cross-Attention) and show that it can efficiently survey a network’s representations and identify strong performing adapter models for a downstream task. 
During training, InCA enables training numerous adapters efficiently and in parallel, isolated from the frozen base model. On the ViT-L/16 architecture, our experiments show that a single adapter, 1.3\% of the full model, is able to reach full fine-tuning accuracy on average across 11 challenging downstream classification tasks. 
Compared with other forms of parameter-efficient adaptation, the isolated nature of the InCA adaptation is computationally desirable for large-scale models. For instance, we adapt ViT-G/14 (1.8B+ parameters) quickly with 20+ adapters in parallel on a single V100 GPU (76\% GPU memory reduction) and exhaustively identify its most useful representations.
We further demonstrate how the adapters learned by InCA can be incrementally modified or combined for flexible learning scenarios and our approach achieves state of the art performance on the ImageNet-to-Sketch multi-task benchmark.
\end{abstract}

\section{Introduction}
Foundation models promise to achieve top performance with minimal adaptation on any downstream task.
In the realm of language, the data and the hypothesis spaces are shared, and many tasks can be unified into a homogeneous representation. Visual inference domains, on the other hand, can be highly heterogeneous and possibly antagonistic. 
For instance, the hypothesis space for pose estimation is geometric, whereas for scene classification it is semantic and even domains that appear homogeneous, such as image classification into the 10 CIFAR classes, can trigger interference in the trained model in the presence of modest perturbations of the image statistics. 

Antagonistic domains may interfere within the activations of the latter layers, which cannot be simultaneously minimal and sufficient for all domains. However, information about a dissimilar domain {\em may be present} in earlier layers, and certainly in the input data which is trivially sufficient for any task. 
Indeed, as opposed to just operating with the final model's representations, the typical approach of addressing domain variability in transfer learning is by applying full fine-tuning of the model on new data. By optimizing all of the model's parameters, each of the model representations can be potentially harnessed as a starting point for the adaptation. 

While accurate and robust to a variety of domains, full fine-tuning of large-scale models entails sweeping computational and storage costs. Further, the resultant model can only function for its dedicated task, not allowing for shared computation and parallel execution of potential future tasks. To tackle the problem of efficient, versatile, and modular adaptation we introduce Introspective Cross-Attention (\name). \name operates on the base-model by attaching isolated shallow adapter modules utilizing any of the activation maps for downstream tasks.

\begin{figure}[t]
\centering
\hspace{0.7em}
  \includegraphics[width=0.45\textwidth]{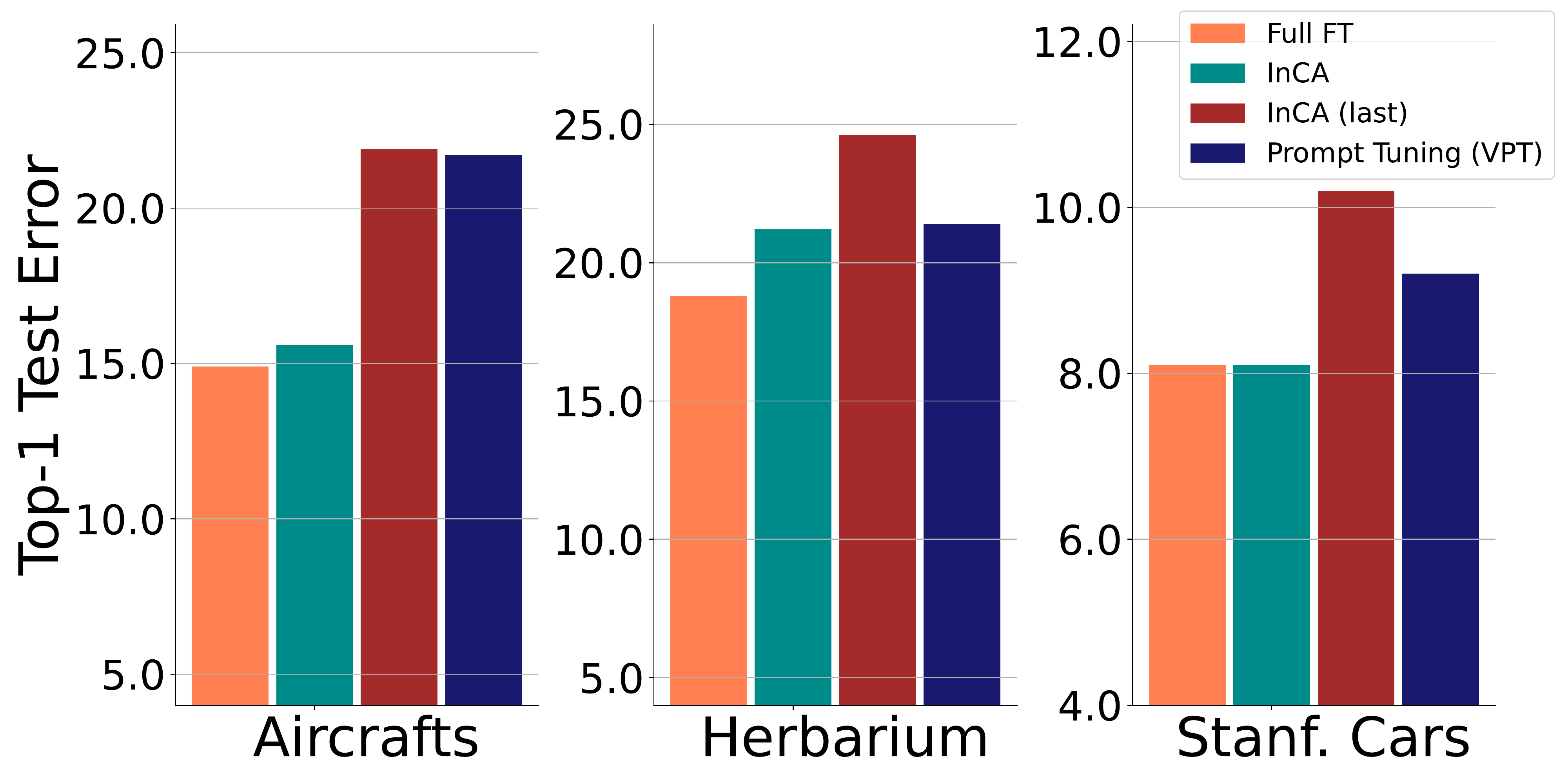}
  \vspace{1em}
  \includegraphics[width=0.46 \textwidth]{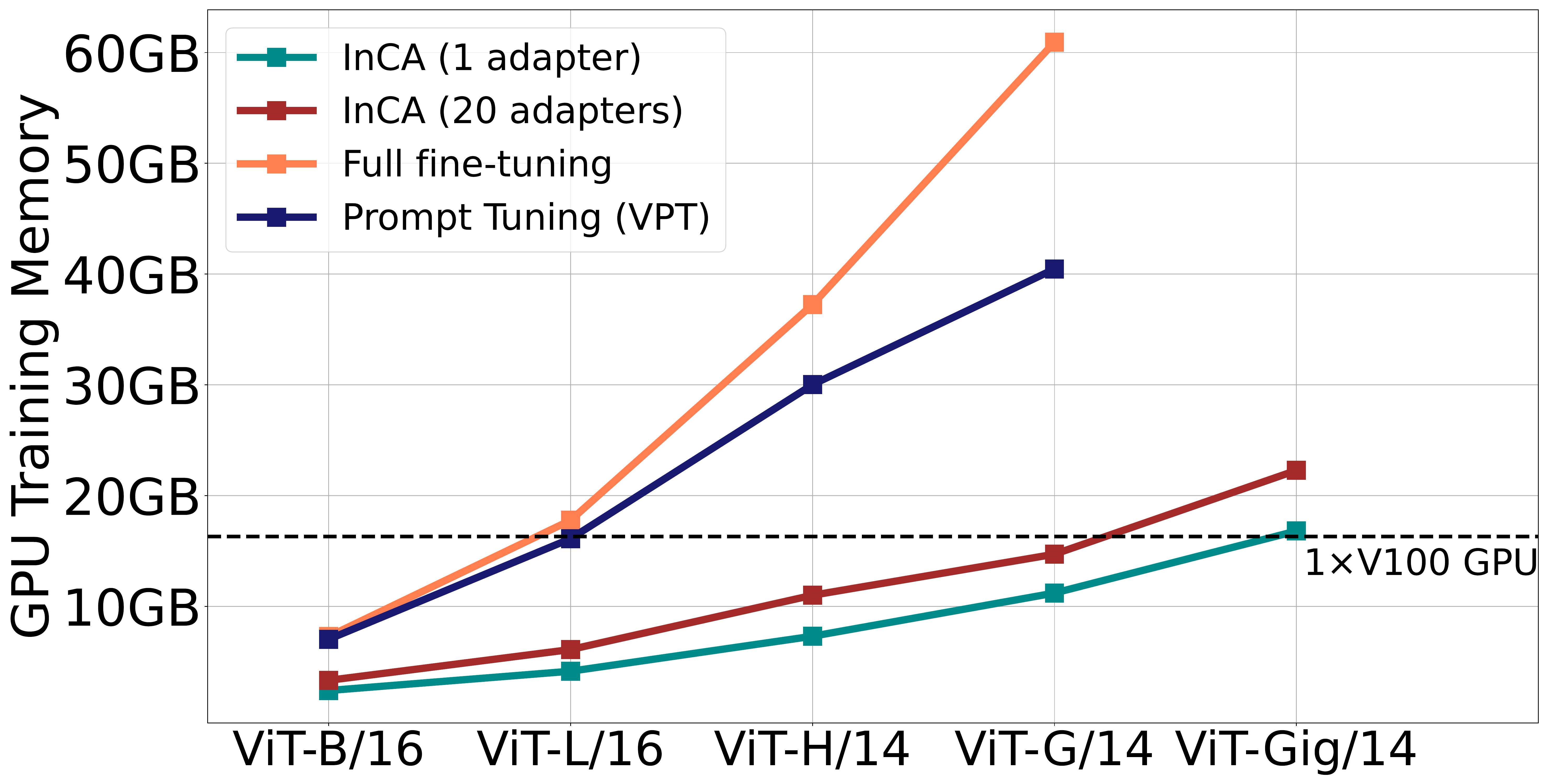} \\
   \caption{\textbf{(Left) Top-1 Test Error} for fine-grained classification transfer learning tasks evaluated with the ViT-L/16 architecture. \name~performs comparable to full fine-tuning on each challenging dataset.
  \textbf{(Right) Max GPU Memory} usage during training for different adaptation approaches and model sizes.
  }\label{fig:scaling}
\end{figure}

\begin{figure*}[t]
\centering
    \includegraphics[width=13.5cm]{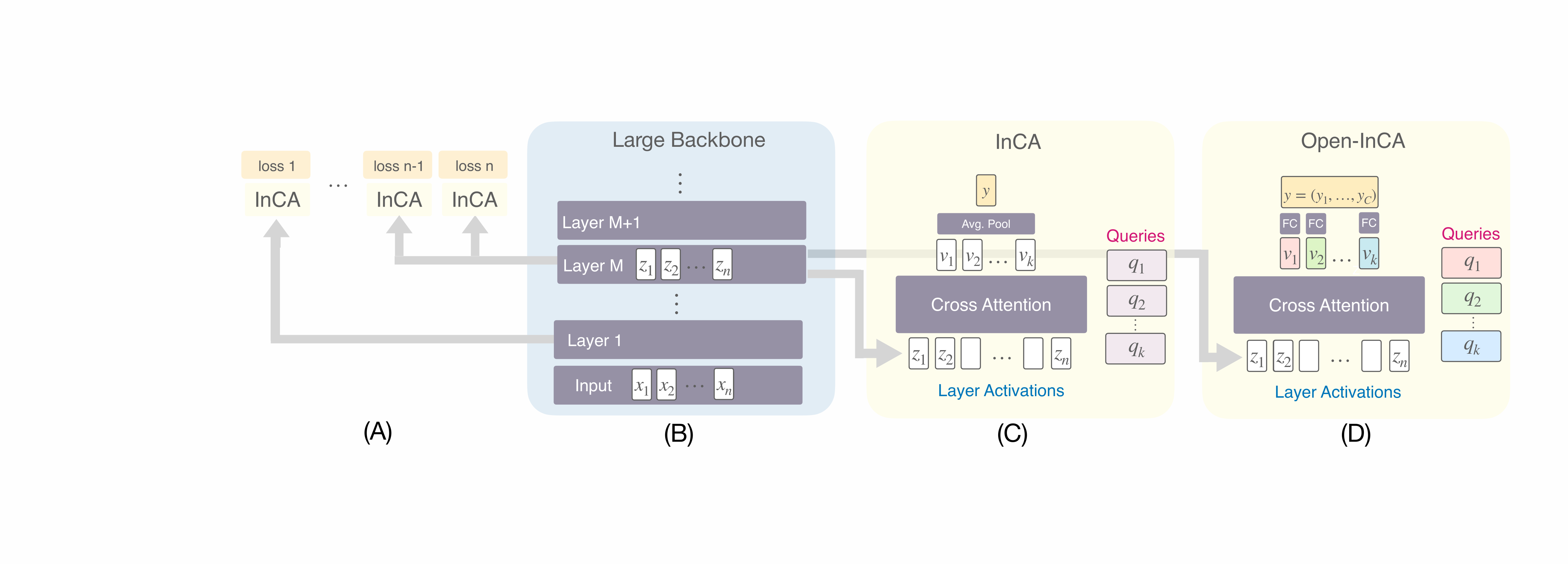}
    \caption{\textbf{\name Adaptation} In (B),
        intermediate activation maps are extracted from a pretrained backbone during forward pass. Each activation map is passed to a lightweight InCA adapter (shown in (C)) or Open-InCA adapter (shown in (D)) depending on the task settings. In (A), we illustrate how multiple adapters are trained in parallel and independently, and during inference can be combined or used in parallel. In (C), (D) we present a schema of the InCA and Open-InCA adapters; see Sec. \ref{sec:method} for details.
 }   
    \label{fig:cal_diagram}
\end{figure*}

Since modern architectures are deep and so is the variety of possible downstream tasks, iterating over all the possible candidate activations of a model is prohibitively expensive. Instead in \name we search for useful model representations exhaustively and in parallel by training numerous isolated adapters attached to different activations. When using \name, each adapter is light and the base-model is fixed and does not require any backpropagation. This makes \name computationally efficient in terms of GPU memory which is crucial for scaling to large models. Further, the shallow \name adapter networks simplify the training dynamics as compared with existing approaches which speeds up training considerably and makes the optimization straightforward and robust (see Appendix \ref{sec:eff_results}).

In detail, during parallel training of \name, a set of adapters sharing the same architecture are trained simultaneously and independently on a downstream task. Each adapter accepts an assigned activation and does not feed back to the backbone, leaving the backbone execution unaltered during both training and inference (Fig.~\ref{fig:cal_diagram}). At inference, the learned adapters may be combined or a best performing adapter can be selected for downstream prediction. 
The \name~adapter architecture is simple, consisting of a single cross-attention module followed by a linear classifier for the downstream task. Despite the simplicity of the adapter, we observe that a single top-performing adapter trained for a downstream task is capable of achieving strong performance across a variety of different architectures and pre-trainings when tested on a diverse set of visual domains. Because our approach does not modify the execution of the model or any of the pre-trained layers as done in existing parameter-efficient approaches \cite{hu2021lora,jia2022visual,houlsby2019parameter}, our method can be automatically applied to \emph{any} architecture without the hassle of re-implementation and architecture specific modifications.
In particular, we present results of \name for ViT~\cite{dosovitskiy2020image}, SWIN~\cite{liu2021swin}, and CNN architectures \cite{liu2022convnext, xie2017aggregated} on a suite of fine-grained recognition tasks. 

Since the adapters learned in \name are shallow and small (1.3\% of the parameters on ViT-L/16), the strong performance of a single adapter implies that many pre-trained models \emph{already contain} strong representations for a diverse set of downstream tasks. Instead, previous approaches like linear probing fall short in using the representations, not because the task can not be solved using the existing model, but rather because of not having the right ``extraction capacity'' of cross-attention; we explore this systematically in Sec. \ref{sec:results} and present a theoretical proof for the advantage of the cross-attention layer in Appendix \ref{sec:theory}.
For challenging datasets, using intermediate
representations as opposed to only adapting the last layer's
representation is key to making \name~a versatile method that
closes the gap with full fine-tuning on diverse tasks (See Fig.~\ref{fig:scaling}).

A byproduct of the exhaustive approach of \name adaptation is a signature describing the performance of the internal representations of a network on different downstream tasks. This renders \name as a powerful tool for understanding the underlying representations of pre-trained models and task space \cite{Achille_2019_ICCV} (See Sec. \ref{sec:analysis}, Appendix \ref{sec:intermediate_rep}). Curiously, we observe that in certain downstream datasets, different pre-trained models hold similar performance signatures, even when the pre-trained models use different architectures or pre-training augmentations.

The isolated training of the adapters means no backpropagation through the pre-trained model is taking place. This significantly reduces the correlation between adaptation cost and the pre-trained model size, which makes it possible to leverage very large architectures even with modest compute resources. 
Fig. \ref{fig:scaling} shows that one V100 GPU can train \name~with 40 adapters using an architecture as big as ViT-G/14. In contrast, existing parameter efficient approaches that backpropagate through the architecture exhaust GPU memory with any model larger than ViT-B/16.

Our contributions are summarized as
\begin{itemize}[itemsep=0pt]
\item We introduce \name, a method that enables versatile downstream adaptation by surveying the internal network representations of any frozen model with lightweight cross-attention based modules as an alternative to full fine-tuning. On ViT-L/16, \name matches full fine-tuning performance and reaches within $0.7\%$ accuracy for a SWIN-L backbone on average over 11 diverse downstream domains.
\item We demonstrate how the modular adapter architecture of \name enables flexible learning and inference scenarios and present the Open-InCA adapter. With our approach we unlock powerful pre-trained models for reusable and parallel multi-task inference, and class-incremental learning.
\item On the efficiency front, \name scales to massive scale architectures under typical computation budgets while its implementation can be automatically applied to any new model. For example, training ViT-G/14 using \name results in 76\% GPU memory reduction as compared with full fine-tuning.  Further, \name is easy to optimize and highly parameter efficient (See Appendix \ref{sec:eff_results}). 

\end{itemize}

The rest of the paper is organized as follows: In Sec. \ref{sec:back} we review related work, and in Sec. \ref{sec:method} we present our approach. We empirically evaluate \name on a wide set of visual recognition tasks in Sec. \ref{sec:results}. Lastly we provide analysis of intermediate representation signatures (Sec. \ref{sec:analysis}) followed by discussion (Sec. \ref{sec:discussion}). Additional results and analysis including Open-InCA are presented in the Appendix.

\section{Related works}\label{sec:back}
\paragraph{Transfer learning} Transfer learning in deep neural networks aims at endowing existing models with new ``downstream'' knowledge. 
The de-facto approach for transfer learning in deep learning modifies an existing pre-trained network by applying full, partial or linear fine-tuning on the model \cite{kolesnikov2020big,shen2021partial,li2016revisiting}. Depending on the the variability between the source and downstream task domains, different approaches aim at delineating different transfer domains \cite{Achille_2019_ICCV,Pandy_2022_CVPR,you2020co}, extending coverage of transfer via suitable pre-trainings such as meta-learning \cite{lee2019meta,finn2017model, ilharco2022patching}, or by selecting and combining from a set of expert models \cite{deshpande2021linearized,gontijo2021no}. More broadly the field of representation learning focuses on learning transferable representations \cite{kornblith2019better} via different pre-training strategies that can be applied for downstream tasks \cite{chen2020simple}. 
\paragraph{Efficient adaptation methods}
In recent times, the top-performing pre-trained model architectures are becoming considerably larger \cite{radford2021learning,liu2022convnext}, and we arrive at a crossroad as full fine-tuning of such large models is becoming out of reach in many practical settings. Recent works address the storage costs associated with large model transfer by proposing parameter efficient transfer approaches as opposed to storing all of the fine-tuned model parameters \cite{he2021towards}. These approaches achieve parameter efficiency by training a subset of existing parameters \cite{zaken2021bitfit,li2016revisiting}, inserting new weight parameters into existing modules \cite{hu2021lora,houlsby2019parameter,NEURIPS2020_bc573864,devaguptapu2023deltapatching} or via additional learnable activations or inputs, known as prompts \cite{jia2022visual,sandler2022fine, lester2021power}.  Compared with existing work \cite{zhang2020side,yang2022rep}, \name is also parameter efficient, yet we place special emphasis on compute and optimization efficiency, especially in the large-scale model setting. Additional lines of work study learning via selective tuning, enabling multi-domain transfer \cite{wallingford2022task,guo2019spottune}.
\paragraph{Feature extraction with attention}
Self and cross-attention mechanisms aggregate information from a set of feature tokens \cite{vaswani2017attention,lee2019set,li2021align} and create representations based on relative inter-feature importance as computed by the attention mechanism \cite{chorowski2015attention,vaswani2017attention}. Self-attention plays a key in the transformer architecture \cite{vaswani2017attention} enabling non-local information aggregation.  In settings where the number of inputs and desired outputs differ, cross-attention enables flexible aggregation based on a pair of query and key feature sets.
When using cross-attention, one can cross-attend between different sets of activations \cite{li2021align,devlin2018bert} or between a set of activations and learnable latent parameters \cite{jaegle2021perceiver,carion2020end,zhu2020deformable}. 
In our settings, the adapter architecture applies cross-attention on extracted representations from a pre-trained network, inspired by the cross-attention module of Perceiver \cite{jaegle2021perceiver}. However, we train multiple adapters in parallel and avoid the iterative re-sampling architecture present in their work. 
More generally, cross-attention layers have been vital in many existing object detection and multi-modal systems that fuse activations
\cite{li2021align,pmlr-v162-borgeaud22a}, or apply cross-attention with learnable latents \cite{alayrac2022flamingo,jaegle2021perceiver,yu2022coca,carion2020end,zhu2020deformable}.
\paragraph{Learning with intermediate features} The re-use of intermediate representations in deep learning is vast and spans from works on interpretability \cite{zeiler2014visualizing}, to state of the art approaches in object detection that harness intermediate layers for multi-resolution feature pyramids \cite{lin2017feature,vasconcelos2022proper,ghiasi2019fpn} and segmentation \cite{hariharan2015hypercolumns,iglovikov2018ternausnet}. For ConvNets, the work of \cite{alain2016understanding} studies classification utilizing intermediate network representations and the authors observe a decrease in accuracy when probing earlier layers.  
\section{Method}\label{sec:method}
We introduce InCA, a lightweight and modular transfer learning alternative to full fine-tuning, that avoids backpropagation through the base-model. Let $f(x) = g_n \circ g_{n-1} \circ \ldots g_1(x)$ be a pre-trained feed-forward neural network of $n$ layers, with $g_j(\cdot)$ corresponding to the $j$-th layer of the network.  We denote the activation computed by $g_j$ as $f_j(x) = g_j \circ g_{j-1} \circ \ldots g_1(x)$. During network inference, a ``forward'' computation processes and computes each $f_j(x)$ activation to arrive to the network's final prediction $f(x)$. During standard training, all of the intermediate activations $\{f_1(x), \dots f_{n-1}(x), f_{n}(x)\}$ are held in GPU memory and are used to compute gradients to update the model. For large models, this incurs large computational and GPU memory costs \cite{rajbhandari2020zero} which limits using the best and largest available pre-trained models under typical computation budgets.

Instead, we attach a set of isolated ``models'' to the pre-trained model $f$ at selected activations $f_{j_k}$ and pass them as input to a set of lightweight and shallow networks $h_k(a)$ with separate parameters and losses. With this, we can train a set of heterogeneous adapters $h_k(a)$ in parallel, while computing inference of the pre-trained model $f$ only once during each update (see Fig. \ref{fig:cal_diagram}). For a set of adapters $h_k(a)$ that take as input intermediate activations from $\{f_{j_{k}}\}$ training follows as:
\vspace{-0.3cm}
\begin{enumerate}
\item  Single inference of $f$ through a data batch $x$ which computes $f(x)$ and selected activations $\{f_{j_k}(x) \}.$ \footnote{We use a callback (though Torch's $\texttt{register\_forward\_hook()}$  or TensorFlow's $\texttt{get\_layer().output}$) to cache the activations of the relevant layers which become leafs of the computational graph}
\item  Calculate the batch predictions and losses for each adapter $h_k$, $\ell_k = \ell(h_k(f_{j_{k}}(x)), y).$
\item  Computing $\ell_{\Sigma} = \sum \ell_k$ and applying automatic differentiation then efficiently resolves the gradient and updates of each $h_k$ automatically as desired.
\end{enumerate}
By avoiding backpropagation through the pre-trained $f$ we decouple the majority of the training costs from depending on the size of the base model $f$ and instead the costs correlate with the much smaller adapter set $\{h_k\}$.
Below we demonstrate that even a simple cross-attention module for $h_k$ makes the overall adaptation sufficiently expressive yet highly efficient.

\paragraph{InCA adapter} 
After extraction of the layer representation $f_k(x)$, we have access to a high-dimensional activation map at our disposal.
To predict a target label $\hat{y}$ from the high-dimensional $f_k(x)$, the typical approach is to apply dimension reduction such as averaging (avgpool) or computing maximum values over a subset of the dimensions and then applying a linear classification head:
\[
\hat{y} = \operatorname{head} \circ \operatorname{avg-pool} \circ f_m(x).
\]
Nonetheless, this simple aggregation approach leads to loss of information which we observe empirically in Sec. \ref{sec:results} and theoretically analyze in Appendix \ref{sec:theory}. Instead, we use a cross-attention module to intelligently aggregate information from the entire large-dimensional activation map $f_k(x)$  into a fixed-dimensional representation based on a number of cross-attention queries. 
Specifically, for standard downstream adaptation, given an intermediate feature map $\mathbf{z} = [z^1, \ldots, z^T] = f_k(x)$ with $T$ tokens or channels  we use the following adapter architecture
\begin{align*}
v_\text{cross}(\mathbf{z})_{[1:m]}  &:= \operatorname{cross-attn}_\theta([z^1, \ldots, z^T], [q_1, \dots, q_m])\\
\inca_\theta(\mathbf{z}) &:=
\operatorname{head}_\theta \circ \operatorname{norm} \,  (\operatorname{avg-pool} (v_\text{cross}(\mathbf{z})_{[1:m]})).
\end{align*}
Note that the query tokens $[q_1, \dots q_m]$ are optimized along with $\theta$. The multi-head cross-attention 
layer outputs $v_\text{cross}$ is produced by surveying the feature map $f_k(x)$ with the query tokens $[q_1, \dots q_m]$. Then, the classification output $\hat{y} = \inca_\theta(\mathbf{z})$ is obtained through averaging the cross-attention outputs (if $m>1$) followed by a fully-connected classification head after normalizing with LayerNorm \cite{ba2016layer}. Based on our experiments, using a single query token $q$ ($m=1$) achieves strong performance and is computationally efficient and we report results with $m=1$ unless otherwise stated. 
\par
For more flexible inference such as in the settings of continual and class-incremental learning tasks, we present a modular version of InCA that disentangles the representations learned between different classes, which we refer to as ``Open-InCA''.
For a $c$-way classification task, define separate queries $[q_1, \dots q_c]$ for each class to compute representations separately,
\begin{align*}
&[v^1_\text{cross}(\mathbf{z}), \dots , v^c_\text{cross}(\mathbf{z})] := \operatorname{cross-attn}_\theta([z^1, \ldots, z^T], [q_1, \dots , q_c])\\
&\openinca_\theta(\mathbf{z}) := \operatorname{diag-head}_\theta \circ \operatorname{norm}   ([v^1_\text{cross}(\mathbf{z}), \dots , v^c_\text{cross}(\mathbf{z})])
\end{align*}
Above, $\operatorname{diag-head}_{\theta}$ is a linear operator layer that operates on a \emph{matrix} input $[a_1, \dots , a_c]$ ``diagonally''. Given a  weight parameter $W$, the operator is defined as the column-wise dot product,
\begin{align*}
    \operatorname{diag-head}_{\theta}([a_1, \dots a_c]) = \big[\langle W_1, a_1 \rangle, \dots \langle W_c, a_c \rangle \big].
\end{align*}

\paragraph{Open-InCA composition}
In the Open-InCA adapter architecture, unique queries $[q_1, \dots, q_c ]$ are defined for each class along with $\operatorname{diag-head}_{\theta}$ that independently processes each coordinate prediction. Both $\operatorname{diag-head}$, LayerNorm and the $\operatorname{cross-attn}$ module in $\openinca$ operate on each input $q_i$ independently which separates the representation learned for each class and enables isolating each adapter output coordinate as
\begin{align*}\label{eq:open_inca_decompose}
\openinca &(\mathbf{z})_i = \langle W_i,\operatorname{norm}(\operatorname{cross-attn}_\theta([z^1, \ldots, z^T], [q_i])) \rangle.
\end{align*}
Above, $W_i$ corresponds to the $i$-th column of $\operatorname{diag-head}$ weight. As a result $\openinca$ enables class-level modularity with the capabilities of new class insertion, deletion and isolated class updates without regression.
For example, deleting class $i$ from the $\openinca$ architecture amounts to simply dropping the query and head parameters $q_i$ and $W_i$ for that coordinate.
In the setting of class-incremental learning (CIL) different query-head pairs from $\openinca$ can be combined together, as long as the parameters of the $\operatorname{norm}$ and $\operatorname{cross-attn}$ remain the same. In practice, this leads to the notion of training $\openinca$ with fixed norm and cross-attention weight parameters, in what we refer to as ``\emph{query-only-training}''. In query-only-training, the learning of a new class corresponds to learning just 2, $d$ dimensional parameters per-class and adapter, where $d$ is the token dimension. Nonetheless, when using pre-trained $\openinca$ layer parameters, ``query-only-training'' performs within the accuracy of \name on many datasets. In Appendix \ref{sec:continual_learning} we compare results of $\inca$, $\openinca$ and query-only-training in class-incremental learning (CIL). In Tab. \ref{tab:open_closed} of the Appendix we observe that even learning just the query and head parameters is capable of harnessing the large dimensional representation maps $f_k(x)$.
\paragraph{Layer branching candidate selection} The cross-attention adapters can be applied in parallel over any intermediate layer of the network and we observe that the performance of many tasks hinges on identifying the right intermediate layer to use for that task.
When considering intermediate activations $f_k(x)$, we observe that
\vspace{-0.3cm}
\begin{itemize}
    \item Using activations such that $f_j(x)$ is directly computed from a residual connection yields better adapter accuracy. This reflects that network representations are refined through each residual block. 
    \item The middle and later layers of the network provide stronger input representations for the adapter. This is likely since the representations of the early layers do not have discriminative enough features to be used directly for high-level tasks. 
\end{itemize}
\paragraph{Two-Stage training}
In settings where the base-model forward-propagation during \name training is too constraining, one may conduct training in two stages. In the first stage, save the activations that serve the input for the adapter for the entire training set, by running the base-model inference for a single epoch. After saving, the second stage proceeds by training the adapters for $T$ epochs with loaded activations. Suppose the per-epoch cost of the pre-trained model forward-propagation is $C_{PT}$ and the per-epoch cost of adapter optimization is $C_A$, then two-stage training reduces the time of training from $O((C_{PT} + C_A) \times T)$ to $O(C_A \times T + C_{PT})$, where $C_{PT} \gg C_A$.
With two-stage training, we are able to reduce a 30-epoch adapter training job to 30 seconds for a cached Stanf. Cars dataset ($\sim$8,000 training samples). We speculate that further optimization can reduce training costs to ``real-time'', enabling an array of user-interactive applications.
\section{Experiments}\label{sec:results}
\paragraph{Datasets}
In our experiments, we measure the capabilities of \name on a diverse set of 11 fine-grained datasets consisting of: CUB-200 \cite{WahCUB_200_2011}, Aircrafts, \cite{maji2013fine}, Stanford Cars \cite{KrauseStarkDengFei-Fei_3DRR2013}, Stanford Dogs \cite{KhoslaYaoJayadevaprakashFeiFei_FGVC2011}, Oxford Flowers 102 \cite{Nilsback08}, MIT-67 \cite{quattoni2009recognizing}, Oxford Pets \cite{parkhi12a}, Describable Textures (DTD) \cite{cimpoi14describing}, European Flood \cite{barz2019enhancing}, FGVC Herbarium \cite{herbarium}, and EuroSAT Land Use dataset \cite{helber2019eurosat}. In Table \ref{tab:multitask} we explore \name~in the settings of multi-task learning and evaluate it on the ImageNet-to-Sketch benchmark that is comprised of 5 datasets: WikiArt \cite{DBLP:journals/corr/SalehE15}, Oxford Flowers \cite{parkhi12a}, Sketch \cite{wang2019learning}, Stanford Cars \cite{KrauseStarkDengFei-Fei_3DRR2013}, and CUB-200 \cite{WahCUB_200_2011}.
\begin{table*}[t]
\centering
\renewcommand{\arraystretch}{1}
\setlength{\tabcolsep}{1.5pt}
  \resizebox{1\textwidth}{!}{
  \begin{tabular}{c|c|cccccccccc}
  \toprule
 ~ & \multicolumn{11}{c}{\textbf{Top-1 Test Error, \underline{ViT-L/16}}} \\
 \midrule
                                  Dataset & Full FT & InCA & InCA (last) &   In. LP &  LP & In. MLP-3 & MLP-3 & VPT \cite{jia2022visual}&       LoRA \cite{hu2021lora} &   ~AdaLN\cite{li2016revisiting} & ~BitFit \cite{zaken2021bitfit} \\
\midrule
    CUB-200 &              9.1 &              \textbf{8.7} &         9.4 &        16.2 &        16.2 &         13.9 &        13.9 &       10.4 &       12.7 &        15.6 &        15.4 \\
       DTD &             18.2 &             \textbf{17.2} &        18.4 &        18.9 &        20.6 &         17.4 &        20.1 &       21.4 &       19.4 &        22.2 &        21.9 \\
    Flood Depth &             18.9 &             \textbf{17.1} &        19.6 &        17.8 &        22.8 &         17.6 &        20.1 &       19.0 &       19.6 &        18.7 &        18.7 \\
    EuroSAT &              1.0 &              1.2 &         1.9 &         2.1 &         3.7 &          1.5 &         2.5 &        1.1 &        \textbf{0.9} &         1.5 &         1.4 \\
    Aircrafts &             14.9 &             \textbf{15.6} &        21.9 &        50.6 &        67.4 &         36.8 &        47.4 &       21.7 &       16.6 &        28.5 &        27.2 \\
    Herbarium &             18.8 &             \textbf{21.1} &        24.6 &        32.6 &        39.8 &         29.5 &        36.4 &       21.4 &       19.2 &        27.9 &        28.3 \\
    MIT-67 &             10.4 &              \textbf{9.0} &         \textbf{9.0} &         9.7 &        10.5 &         10.1 &        11.2 &       14.8 &       14.8 &        15.1 &        15.1 \\
    Oxford Flowers &              0.6 &              \textbf{0.3} &         0.4 &         0.6 &         1.1 &          0.5 &         0.7 &        2.2 &        4.0 &         7.0 &         7.2 \\
    Oxford Pets &              4.2 &              \textbf{4.0} &         4.2 &         6.1 &         6.4 &          5.3 &         5.5 &        6.9 &        4.3 &         5.5 &         5.3 \\
    Stanf. Cars &              8.1 &              \textbf{7.7} &        10.2 &        29.2 &        47.2 &         20.8 &        31.4 &        9.2 &        8.4 &        16.0 &        14.7 \\
    Stanf. Dogs &              5.9 &              5.4 &         5.8 &         5.3 &         5.3 &          5.7 &         5.7 &        7.3 &        4.3 &         3.8 &         \textbf{3.7} \\
                                    \midrule
\makecell{Mean Top-1 Test Error \\ (Max. gap to Full FT)} &          \makecell{10.0  (0.0)} &  ~\textbf{9.8 (-2.3)} &  11.5 (-7.0) & 17.2 (-35.7) & 21.9 (-52.5) &  14.5 (-21.9) 
& 17.7 (-32.5) & 12.3 (-6.8) & 11.3 (-4.4) & 14.7 (-13.6) & 14.4 (-12.3) \\
\midrule
\% Trainable param. &   100\% &  1.3\% &  1.3\% & 0.1\% & 0.1\% &  2.8\% & 2.8\% & 0.8\% & 2.4\% & 0.1\% & 0.1\% \\
\midrule 
No backbone backprop. &  \xmark &  \checkmark &  \checkmark & \checkmark & \checkmark &  \checkmark & \checkmark & \xmark & \xmark & \xmark & \xmark \\
\bottomrule
\end{tabular}
}
  \caption{\textbf{Fine-grained Classification Top-1 Test Error (ViT-L/16)} We compare \name to full fine-tuning (Full FT) along with other adaptation approaches for downstream learning. For each method we summarize the maximum gap in performance compared with the full fine-tuning paragon. In addition, we report the parameter efficiency and whether the method requires backpropagation through the pre-trained model.  The minimum error over the columns excluding Full FT is presented in bold.}
  \label{tab:fine_grained_vit}
\end{table*}

\begin{table*}[b]
    \centering
   \resizebox{1\textwidth}{!}{
\begin{tabular}{lllllllll}
\toprule
~ & ~&~ & \multicolumn{5}{c}{Mean Top-1 Test Error (Max. gap to full FT)} \\
\toprule
         
              Category & Architecture &  Pretraining data & Full FT & InCA & InCA (last) & inter. LP & LP & Model size \\
\midrule
             ~~~Vanilla   &         ViT-B/16~\cite{dosovitskiy2020image} &               In21K &         13.0 (0) & 15.9 (-7.6) & 17.5 (-16.4) &               23.9 (-32.4) &   24.3 (-32.4) &               86.5M              \\
             Transformer  &         ViT-B/16~\cite{li2021align}  &       ALBEF (CC14M) &              13.8 (0) &        13.5 (-4.2) &         14.8 (-9.3) &                       24.7 (-42.6) &           25.8 (-42.6) &           85.9M                   \\
              &         ViT-L/16~\cite{touvron2022deit}  &        In21K (DeiT) &         10.0 (0) & 9.8 (-2.3) &        11.5 (-7) &        17.2 (-35.7) &     21.9 (-52.5) &      304.3M      \\
                          &   ViT-L/16 @384~\cite{touvron2022deit}  &      In21K (DeiT)  &              -$^\dagger$ &         9.2 (-0.6$^\dagger$) &          11.7 (-9.1$^\dagger$) &                        17.3 (-38.1$^\dagger$) &            22.0 (-54.4$^\dagger$) &      304.7M                        \\
              & CLIP-ViT-L/14@336~\cite{radford2021learning} &  400M Im-Text &               -$^\dagger$ &         9.2 &          10.6 &                        19.6 &            21.8 &                  304.2M \\
              &   ViT-H/14 \cite{dosovitskiy2020image} &      2B Im-Text   &              -$^\dagger$ &         9.4  &          10.4  &                    14.0 &            15.2 &      632.8M                        \\
               &   ViT-G/14 \cite{cherti2023reproducible} &      2B Im-Text    &              -$^\dagger$ &         9.6  &          10.4  &                  15.3   &            16.8 &      1884.9M                        \\
              \midrule
         Hier. Transformer &           SWIN-L~\cite{liu2021swin} &               In21K &          9.3 (0) &   10 (-3.6) &  12.4 (-9.5) &               15.8 (-31.3) &   18.3 (-40.5) &              196.5M               \\
         \midrule
Convolutional &       ConvNext-B~\cite{liu2022convnext} &               In21K &          9.4 (0) & 10.7 (-7.4) & 12.5 (-12.6) &               19.1 (-44.2) &   19.4 (-44.2) &                  88.5M           \\
  &          ResNext-101~\cite{xie2017aggregated} &     IG-3.5B \cite{mahajan2018exploring}                &         11.4 (0) &   12 (-8.7) & 17.3 (-27.1) &               20.1 (-38.8) &   21.3 (-39.7) &      468.5M                       \\
\bottomrule
\end{tabular}
}\\

    \caption{\textbf{Mean Top-1 Test Error} for transfer learning with a variety of ViT, SWIN, and convolutional networks, including different network scales and pre-training strategies. Averages are reported on the 11 datasets presented in Table \ref{tab:fine_grained_vit}.$^\dagger$ indicates Full FT was avoided due to prohibitive computational costs. For DeiT ViT-L/16 @384 the gap is computed with respect to the 224 pre-training.}
    \label{tab:many_arch}   
\end{table*}

\begin{table*}[b]
  \centering
  \resizebox{1\textwidth}{!}{
  \setlength{\tabcolsep}{1.5pt}{
\begin{tabular}{@{}c|c|ccccccccc@{}}
\bottomrule
 ~ & \multicolumn{10}{c}{\textbf{Top-1 Test Error, \underline{SWIN-L}}} \\
 \midrule
        Dataset      & Full FT &       InCA & InCA (last) &   In. LP &          LP & In. MLP-3 &       MLP-3 &   LoRA\cite{hu2021lora} &   AdaLN\cite{li2016revisiting} & ~BitFit\cite{zaken2021bitfit} \\
            \midrule
                         CUB-200 & 9.0 &  9.1 &   9.6 &  10.2 &  10.6 &    9.7 &   9.7  & 10.0 &   9.1 &   \textbf{8.8} \\
                 DTD &       15.6 & 17.8 &  19.1 &  17.7 &  19.1 &   16.7 &  16.7   & \textbf{15.8} &  16.7 &  17.0 \\
                Flood Depth &       17.6 & \textbf{16.3} &  18.3 &  18.5 &  18.5 &   16.7 &  18.5 & 17.1 &  16.9 &  17.8 \\
                   EuroSAT &  0.7 &  1.5 &   2.4 &   2.7 &   3.7 &    1.6 &   2.2   & \textbf{0.9} &   1.1 &   1.7 \\
                 Aircrafts &       12.2 & \textbf{15.8} &  25.3 &  43.5 &  52.7 &   33.7 &  34.8  & 16.1 &  22.7 &  26.5 \\
                 Herbarium &       14.9 & \textbf{18.2} &  23.0 &  29.2 &  34.0 &   24.9 &  27.6  & 18.4 &  21.2 &  29.7 \\
              MIT-67 &       10.5 & 10.1 &  10.1 &   9.9 &  10.3 &   10.2 &  10.2  & 9.6 &   8.9 &   \textbf{8.5} \\
             Oxford Flowers &  0.5 &  \textbf{0.3} &   0.4 &   0.5 &   0.5 &    0.5 &   0.5  & 0.4 &   0.4 &   0.4 \\
                Oxford Pets &  4.6 &  \textbf{4.7} &   5.5 &   5.0 &   5.5 &    5.5 &   5.5  & 5.2 &   4.8 &   4.9 \\
                Stanf. Cars &  7.3 &  \textbf{8.4} &  15.0 &  29.2 &  39.0 &   22.4 &    26  & 9.6 &  14.2 &  18.4 \\
                Stanf. Dogs &  9.1 &  8.1 &   8.1 &   \textbf{7.1} &   \textbf{7.1} &    9.8 &   9.8  & 11.3 &   9.1 &   9.0 \\
    \midrule
\makecell{Mean Top-1 Test Error \\ (Max. gap to Full FT)}  &           9.3 (0) &  \bf 10.0 (-3.6) &  12.4 (-9.5) & 15.8 (-31.3) & 18.3(-40.5) &  13.8 (-21.5) & 14.7 (-22.6) & 10.4 (-3.9) & 11.4 (-10.5) &   13.0 (-14.8) \\
\midrule
\% Trainable param.$^{\S}$ &   100\% &  3.7\% &  3.7 \% & 0.1\% & 0.1\% &  2.8\% & 2.8\% & 0.8\% & 0.1\% & 0.1\% \\
\midrule 
No backbone backprop. &  \xmark &  \checkmark &  \checkmark & \checkmark & \checkmark &  \checkmark & \checkmark & \xmark & \xmark & \xmark \\
\bottomrule
\end{tabular}
}
}
  \caption{\textbf{Fine-grained Classification Top-1 Test Error (SWIN-L)} We compare \name to full fine-tuning (Full FT) along with other adaptation approaches for downstream learning. For each method we summarize the maximum gap in performance compared with the full fine-tuning paragon. In addition we report the parameter efficiency and whether the method requires backpropagation through the pre-trained model. $\S$ For SWIN-L different activations map sizes leads to different $\%$ of trainable parameters and we report the maximum for each method.  The minimum error over the columns excluding Full FT is in bold.}
  \label{tab:fine_grained_swin}
\end{table*}

\paragraph{Baselines}
For downstream transfer experiments, we compare \name adaptation to other adaptation protocols. 1) \emph{paragon}: Full fine-tuning is considered as the paragon as it performs well on a diverse set of datasets but incurs steep computational and parameter costs.
2) We compare \name to other parameter efficient approaches, including 2a) LoRA \cite{hu2021lora} 2b) Visual Prompt Tuning (VPT) \cite{jia2022visual} where we apply the top-performing VPT approach, \emph{VPT-Deep}. 2c) BitFit \cite{zaken2021bitfit} and 2d) AdaLN \cite{li2016revisiting} which is the LayerNorm analogous, AdaBN approach. Note, each approach we name in 2) requires backpropagating through the entire network's activations to update the learnable parameters and leads to a large computational overhead compared with \name. 
3) In addition, we also compare \name with a suite of computationally efficient approaches that avoid backbone backpropagation like \name. These include 3a) Linear Probing (LP),  3b) Intermediate Linear Probing (In. LP) which utilizes the same training procedure as \name but with a LP classifier on the activations. 3c) MLP-3, which is a feed-forward network that consists of probing the base-model with a 3-layer feed-forward network, and 3d) Intermediate MLP-3 (In. MLP-3), the extension of MLP-3 to intermediate layers. 
\paragraph{Training details}
In all of our results (including multi-task settings) we use the same training configuration for \name. We only change the adapter architecture input layer to automatically match the dimension of the base-model activation map. \name~is robust to hyper-parameters and our training schedule is consistent for all runs. This amounts to 30 training epochs, AdamW optimizer~\cite{loshchcilov2019adamw} with 2 learning rates, and cosine annealing learning schedule; we provide the full training details in Appendix \ref{sec:implement}. 
While \name~is efficient enough to operate at larger resolutions, we use 224 image resolution unless stated otherwise. Nonetheless, \name~performance improves at 384 resolution while remaining computationally competitive (see Table \ref{tab:many_arch}).  
\paragraph{Transfer learning on ViT} In Table \ref{tab:fine_grained_vit}, we demonstrate the transfer performance of \name~applied to ViT-L/16. For each dataset we train \name~and extract activations at residual layers of the ViT for the last 12 blocks and output layer.
For all baselines and our method we use the ViT DeiT~pre-training \cite{touvron2022deit} and additionally report ViT-L/16 pre-training results in Appendix Table \ref{tab:many_arch_per_dataset}.  In the table, we compare \name to full fine-tuning as well as applying InCA on the last layer and observe that only \name is capable of achieving good results on challenging datasets such as Aircraft, Stanf. Cars, etc. and closes the maximal gap to full fine-tuning to -2.3\%. The second best adaptation approach is LoRA which achieves a maximum gap of -4.4\% to full fine-tuning, yet at additional training costs. \\
For a single dataset we can train the \name~modules with 2 learning rates in parallel which corresponds to 26 \name~modules with identical architectures attached to 13 activation maps. In this case, the total training costs of \name~on a single dataset correspond to one base-model run. In Appendix \ref{sec:eff_results}, we report the hyper-parameter settings and training cost of \name and current state of the art adaptation method for transformers, VPT~\cite{jia2022visual} which incurs up to $8.7\times$ the training costs of \name with a large hyper-parameter search ($2$ vs. $24$ settings).
\paragraph{Transfer learning on SWIN}
In Table \ref{tab:fine_grained_swin} we present downstream adaptation results for the SWIN-L pre-trained model. \name adaptation is applied to the 3rd and 4th stages of the network residual activations. Because of the heterogeneous activation dimensions of the hierarchical SWIN architecture, the reported adaptation model sizes depend on the activation map used for the selected adapter. \name achieves the smallest maximum gap to full fine-tuning on SWIN while being computationally efficient. As with ViT-L, in SWIN we observe that challenging datasets require using intermediate activation maps for \name, closing the maximal gap from (-9.5\%) to (-3.6\%).
\paragraph{Evaluating \name~on different pre-trained models}
\name can be applied to any feed-forward neural network \emph{without} any implementation changes. We simply specify the intermediate layer names and feature tensor-ordering for any new architecture and \name can be used directly. We note this is in sharp contrast to methods that rely on specific layers such as convolution filters \cite{berriel2021ba2, mallya2018piggyback} or self-attention \cite{hu2021lora, jia2022visual, lester2021power}. We illustrate the architecture versatility of our method in Table \ref{tab:many_arch}. We report the mean and maximum test error gap from full fine-tuning on the 11 fine-grained dataset suite as studied in Table \ref{tab:fine_grained_vit}. We test different architecture families, which include vanilla vision transformers \ie ViTs, SWIN~\cite{liu2021swin}, and modern convolutional networks (ConvNext~\cite{liu2022convnext}, ResNext~\cite{xie2017aggregated}). In addition, we test models pre-trained via different strategies including supervised learning~\cite{dosovitskiy2020image,touvron2022deit} and vision-language objectives~\cite{li2021align,radford2021learning,cherti2023reproducible}. We also test \name at different ViT scales from ViT/B-16 (86M) to ViT/G-14 (1.8B). For \name adaptation, all model sizes were trained on a single V100 GPU with batch size 32, including for the larger input resolutions runs.

\begin{table}
  \centering
  \resizebox{1\textwidth}{!}{
   \begin{tabular}{l| cccccc | rrrrrr}
\toprule
\multicolumn{7}{c}{Top-1 Test Error} & \multicolumn{6}{c}{Adaptation Efficiency} \\
 \midrule
          Method & Avg. & Flowers & WikiArt & Sketch & Cars & CUB-200 &  \multicolumn{2}{c}{\makecell{\# of trainable \\  parameters}} & \multicolumn{2}{c}{\makecell{GPU Memory \\ (training)}} & \multicolumn{2}{c}{\makecell{Inference Time\\ (for all 5 tasks)}} \\
\cmidrule(lr){1-1} \cmidrule(lr){2-13}  
Full fine-tuning & 10.5 &     0.6 &    \bf{14.7} &   14.4 & 10.8 &  12.2 &  \multicolumn{2}{c}{$5\times$} & \multicolumn{2}{c}{$1\times$} & \multicolumn{2}{c}{$5\times$} \\
   Linear probing &           29.8 & 10.9 &    37.2 &   29.3 & 44.5 &    27.9 &  \multicolumn{2}{c}{$0.01\times$} & \multicolumn{2}{c}{$0.17\times$} & \multicolumn{2}{c}{$1.01\times$} \\
BA$^2$~\cite{berriel2021ba2} &         15.9 & 4.3 &    27.7 &   20.7 & 7.9 &    18.8 &  \multicolumn{2}{c}{$1.03\times$} & \multicolumn{2}{c}{$1\times$} & \multicolumn{2}{c}{$5\times$} \\
TAPS~\cite{wallingford2022task} & 10.4 & 0.6 &    15.8 &   \textbf{14.0} & 11.1 &    10.4 &  \multicolumn{2}{c}{$4.12\times$} & \multicolumn{2}{c}{$1.23\times$} & \multicolumn{2}{c}{$5\times$} \\
SpotTune~\cite{guo2019spottune} &   14.3 & 3.7 &    24.2 &   19.8 & \bf{7.6} &    16.0 &  \multicolumn{2}{c}{$5.27\times$} & \multicolumn{2}{c}{$2\times$} & \multicolumn{2}{c}{$7.3\times$} \\
\midrule
            \textbf{\name} &      \textbf{9.8} & \textbf{0.3} &    15.4 &   16.8 & 7.7  &    \textbf{8.8} &  \multicolumn{2}{c}{$0.06\times$} & \multicolumn{2}{c}{$0.51\times$} & \multicolumn{2}{c}{$1.13\times$} \\
\bottomrule
\end{tabular}
}
\vspace{0.2cm}
\caption{\textbf{Multitask Efficiency and Top-1 Test Error} on ``ImageNet-to-Sketch'' benchmark. \name~is the top performing method on average and is parameter efficient. Further, only \name~and linear probing ``share computation'' of the pre-trained model and enable ``one-to-many'' inference execution measured in the ``Inference Time'' column. BA$^2$ is based on ResNet-50 and can not be applied to ViTs. The rest of the methods are based on ViT-L/16. } 
\label{tab:multitask}
\end{table}

\paragraph{Multi-task Experiments}
\name's isolated design is suitable for multi-task inference and a single pre-trained-model can efficiently evaluate a batch of samples on multiple tasks, allowing for ``one-to-many'' inference. We compare \name on the ImageNet-to-Sketch multi-task benchmark in Table \ref{tab:multitask}. All methods except BA$^2$ were trained with a ViT-L/16 model and evaluated with the ImageNet-to-Sketch version of each dataset \cite{mallya2018piggyback}. For BA$^2$~\cite{berriel2021ba2}, we report the adaptation on a ResNet-50~\cite{he2016resnet} backbone, as the BA$^2$ approach requires convolutional filters. Overall, \name~is the top performing method reaching near the paragon on the evaluated datasets. Importantly for multi-task, only \name~and LP enable multi-task inference via ``computation sharing'' of the base model inference.

\paragraph{Learning efficiency}
Isolating the learning from the base-model means \name learns shallow neural networks directly on a downstream task. By avoiding deeply backpropagated gradients through the base model, the adapters receive direct signal which improves the optimization dynamics and speed of training. We compare the number of training steps required to train \name and VPT-Deep and observe that \name can be optimized in 4.5$\times$ fewer epochs than VPT. Here we don't take into account the additional GPU memory costs of optimizing VPT in each step, nor the required hyper-parameter sweeps used in VPT. More detailed efficiency comparison results are given in Appendix \ref{sec:eff_results}.

\section{Analysis \label{sec:analysis}}
\begin{figure}[b]
    \centering
    \includegraphics[width=10cm,height=6cm]{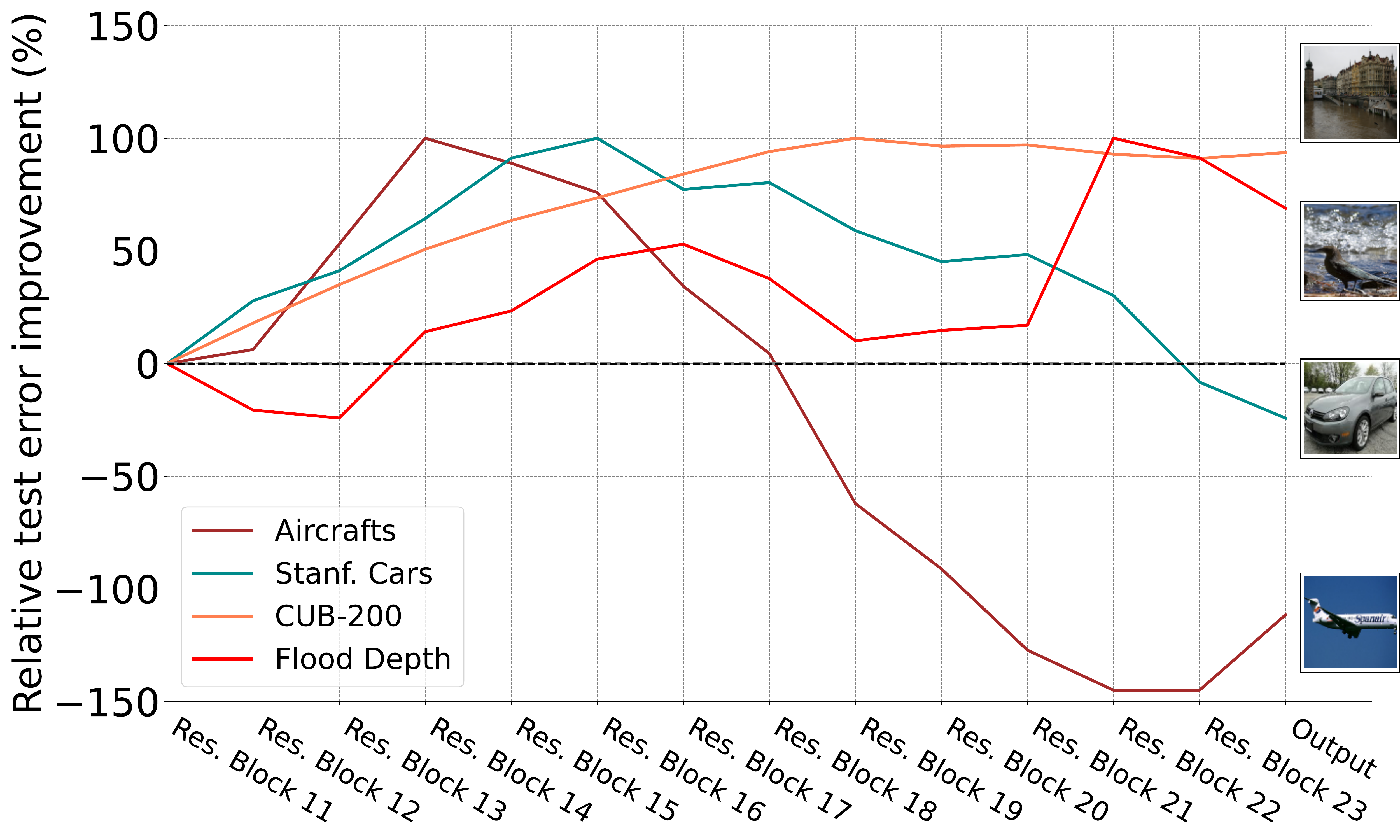}
    \caption{\textbf{\name Layer Performance Signature} Relative test error improvement of \name adapters attached to different intermediate layers. We evaluate \name with ViT-L/16 with adapters at each residual block starting from Block 11.}
    \label{fig:affinity_curve}
\end{figure}

We analyze the results of \name adaptation, focusing on the performance signature of different intermediate representations used as input for the adapter and the relation between the top \name layers with fine-tuning. Further in Appendix \ref{sec:theory}, we provide a theoretical proof motivating the extraction capabilities of cross-attention as it is used in \name.

\paragraph{Intermediate representations}
We consider the intermediate representation signature created by evaluating the accuracy of adapters that utilize different layers. In Figure \ref{fig:affinity_curve}, we review the adapter performance applied to different layer representations. Datasets like CUB-200 and Flood-Depth mostly prefer final representations, whereas for datasets like Aircrafts and Stanf. Cars, the best adaptations use earlier representations with decreasing performance towards the last activations.
Curiously, we observe consistency in layer affinity for certain datasets while using different pre-trainings for the backbone and even when using different architectures (Appendix Fig. \ref{fig:inca_layer_affinity}).

\begin{figure}[ht]
    \centering
    \includegraphics[width=10cm]{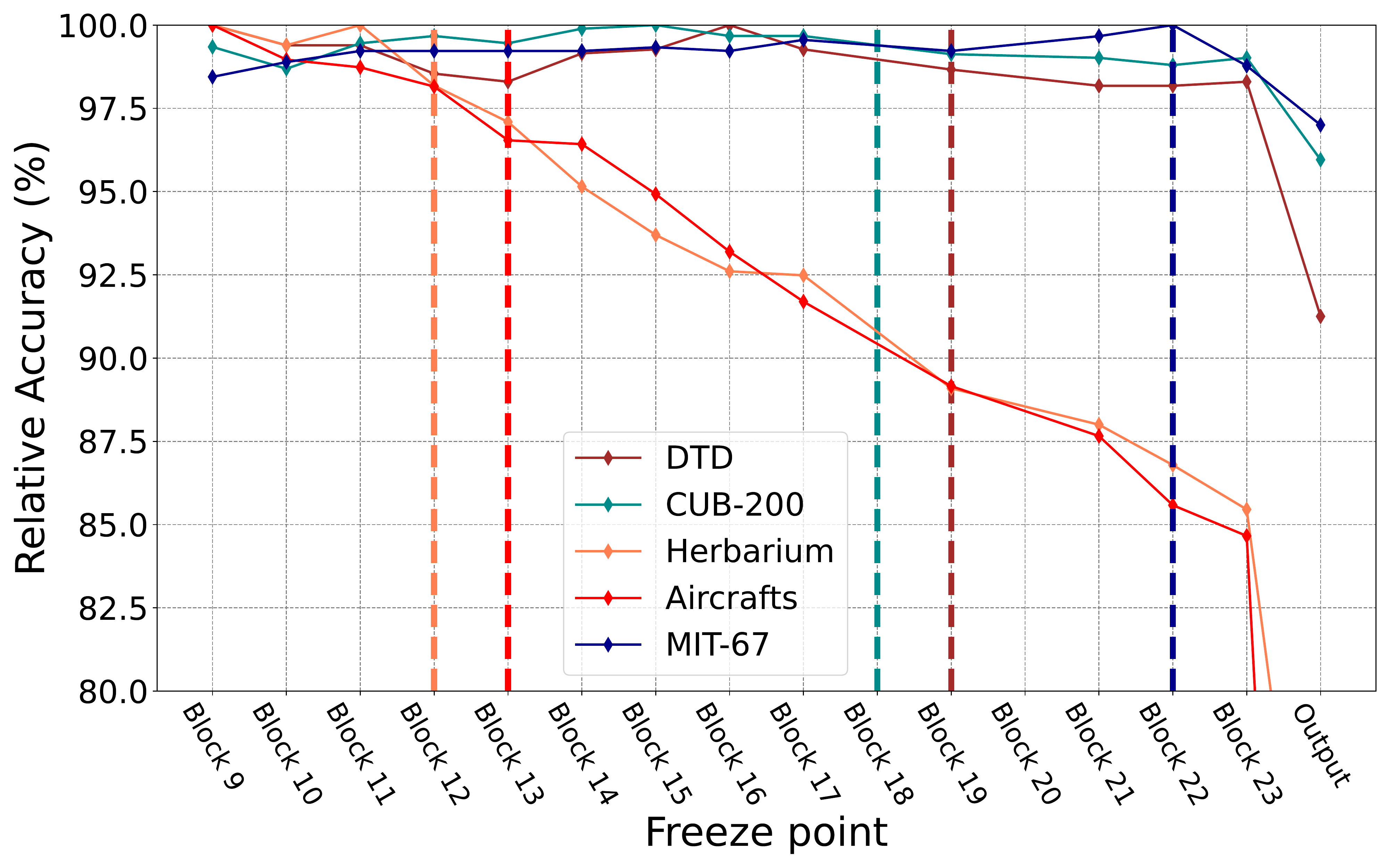} \\
   
    \caption{\textbf{Partial Fine-tuning vs. InCA} Vertical dashed lines indicate the top InCA layer; curves show final test accuracy for different partial tuning training runs. Each mark indicates a run where all of the pre-trained model parameters are trained up to a ``freeze point'' in the network's layers.  Note partial tuning performance saturates in close proximity to the optimal InCA adapter layer. This is aligned with our hypothesis that full fine-tuning attempts to surface \emph{existing representations already in the network}. In that case, performance improves until the tuning approach unlocks the capacity to utilize an existing relevant representation and performance plateaus afterwards. Note here we refer to output layers, \eg,  the adapter at block 19 means the adapter corresponding to the final output of block 19, or the input to block 20.}
\label{fig:partial_tuning}
\end{figure}

\textbf{\name and partial-tuning} 
In Appendix \ref{sec:intermediate_rep} we compare \name with gradually un-freezing the base-model and applying partial fine-tuning on a growing set of layers. 
We run a set of experiments where we fine-tune a pre-trained model starting at different freezing points, this means we optimize all layers of the network after the freezing point location. For each dataset we construct a “partial tuning curve'' where we plot the final test accuracy vs. freezing point (Figure \ref{fig:partial_tuning}). 
Interestingly, we observe a direct correlation between the layer-location of the top \name adapter and the point where the partial tuning curve saturates. 
In particular, the partial tuning test accuracy plateaus (to the tuning of more layers) at around the same layer location as the top performing InCA adapter layer location.
Namely, the point of saturation of the partial tuning curve is where partial-tuning is capable of harnessing the representation found by \name at that layer. This gives further evidence that \emph{``your representations are in the network''} and that fine-tuning surfaces existing representations that can be directly identified by \name. However, \name adaptation operates an order of magnitude more efficiently and scales better to large models.

\section{Discussion} \label{sec:discussion}
In this paper, we present an efficient and effective alternative to full fine-tuning for transfer learning, closing the gap to full fine-tuning on a diverse set of downstream datasets. \name has many benefits: it inherently generalizes to different architectures, efficiently scales to massive models, optimizes effectively, and unlocks modular and flexible adaptation applications including multi-task and incremental learning. Further, through the parallel exhaustive search of \name we are able to better understand the inner representation dynamics of neural networks and construct illuminating ``representation signatures'' of different models and datasets.

{\small
\bibliographystyle{ieee_fullname}
\bibliography{main}

\begin{thebibliography}{10}\itemsep=-1pt

\bibitem{Achille_2019_ICCV}
Alessandro Achille, Michael Lam, Rahul Tewari, Avinash Ravichandran, Subhransu
  Maji, Charless~C. Fowlkes, Stefano Soatto, and Pietro Perona.
\newblock Task2vec: Task embedding for meta-learning.
\newblock In {\em Proceedings of the IEEE/CVF International Conference on
  Computer Vision (ICCV)}, October 2019.

\bibitem{alain2016understanding}
Guillaume Alain and Yoshua Bengio.
\newblock Understanding intermediate layers using linear classifier probes.
\newblock In {\em 5th International Conference on Learning Representations,
  ICLR 2017, Toulon, France, April 24-26, 2017, Workshop Track Proceedings}.
  OpenReview.net, 2017.

\bibitem{alayrac2022flamingo}
Jean{-}Baptiste Alayrac, Jeff Donahue, Pauline Luc, Antoine Miech, Iain Barr,
  Yana Hasson, Karel Lenc, Arthur Mensch, Katherine Millican, Malcolm Reynolds,
  Roman Ring, Eliza Rutherford, Serkan Cabi, Tengda Han, Zhitao Gong, Sina
  Samangooei, Marianne Monteiro, Jacob~L. Menick, Sebastian Borgeaud, Andy
  Brock, Aida Nematzadeh, Sahand Sharifzadeh, Mikolaj Binkowski, Ricardo
  Barreira, Oriol Vinyals, Andrew Zisserman, and Kar{\'{e}}n Simonyan.
\newblock Flamingo: a visual language model for few-shot learning.
\newblock In {\em NeurIPS}, 2022.

\bibitem{ba2016layer}
Jimmy~Lei Ba, Jamie~Ryan Kiros, and Geoffrey~E Hinton.
\newblock Layer normalization.
\newblock {\em arXiv preprint arXiv:1607.06450}, 2016.

\bibitem{barz2019enhancing}
Björn Barz, Kai Schröter, Moritz Münch, Bin Yang, Andrea Unger, Doris
  Dransch, and Joachim Denzler.
\newblock Enhancing flood impact analysis using interactive retrieval of social
  media images.
\newblock {\em Archives of Data Science, Series A (Online First)}, 5(1):A06, 21
  S. online, 2018.

\bibitem{zaken2021bitfit}
Elad Ben~Zaken, Yoav Goldberg, and Shauli Ravfogel.
\newblock {B}it{F}it: Simple parameter-efficient fine-tuning for
  transformer-based masked language-models.
\newblock In {\em Proceedings of the 60th Annual Meeting of the Association for
  Computational Linguistics (Volume 2: Short Papers)}, pages 1--9, Dublin,
  Ireland, May 2022. Association for Computational Linguistics.

\bibitem{berriel2021ba2}
Rodrigo~Ferreira Berriel, St{\'{e}}phane Lathuili{\`{e}}re, Moin Nabi, Tassilo
  Klein, Thiago Oliveira{-}Santos, Nicu Sebe, and Elisa Ricci.
\newblock Budget-aware adapters for multi-domain learning.
\newblock In {\em 2019 {IEEE/CVF} International Conference on Computer Vision,
  {ICCV} 2019, Seoul, Korea (South), October 27 - November 2, 2019}, pages
  382--391. {IEEE}, 2019.

\bibitem{pmlr-v162-borgeaud22a}
Sebastian Borgeaud, Arthur Mensch, Jordan Hoffmann, Trevor Cai, Eliza
  Rutherford, Katie Millican, George~Bm Van Den~Driessche, Jean-Baptiste
  Lespiau, Bogdan Damoc, Aidan Clark, Diego De~Las~Casas, Aurelia Guy, Jacob
  Menick, Roman Ring, Tom Hennigan, Saffron Huang, Loren Maggiore, Chris Jones,
  Albin Cassirer, Andy Brock, Michela Paganini, Geoffrey Irving, Oriol Vinyals,
  Simon Osindero, Karen Simonyan, Jack Rae, Erich Elsen, and Laurent Sifre.
\newblock Improving language models by retrieving from trillions of tokens.
\newblock In Kamalika Chaudhuri, Stefanie Jegelka, Le Song, Csaba Szepesvari,
  Gang Niu, and Sivan Sabato, editors, {\em Proceedings of the 39th
  International Conference on Machine Learning}, volume 162 of {\em Proceedings
  of Machine Learning Research}, pages 2206--2240. PMLR, 17--23 Jul 2022.

\bibitem{carion2020end}
Nicolas Carion, Francisco Massa, Gabriel Synnaeve, Nicolas Usunier, Alexander
  Kirillov, and Sergey Zagoruyko.
\newblock End-to-end object detection with transformers.
\newblock In {\em European conference on computer vision}, pages 213--229.
  Springer, 2020.

\bibitem{chaudhry2018efficient}
Arslan Chaudhry, Marc’Aurelio Ranzato, Marcus Rohrbach, and Mohamed
  Elhoseiny.
\newblock Efficient lifelong learning with a-{GEM}.
\newblock In {\em International Conference on Learning Representations}, 2019.

\bibitem{chen2020simple}
Ting Chen, Simon Kornblith, Mohammad Norouzi, and Geoffrey Hinton.
\newblock A simple framework for contrastive learning of visual
  representations.
\newblock In {\em International conference on machine learning}, pages
  1597--1607. PMLR, 2020.

\bibitem{cherti2023reproducible}
Mehdi Cherti, Romain Beaumont, Ross Wightman, Mitchell Wortsman, Gabriel
  Ilharco, Cade Gordon, Christoph Schuhmann, Ludwig Schmidt, and Jenia Jitsev.
\newblock Reproducible scaling laws for contrastive language-image learning.
\newblock In {\em Proceedings of the IEEE/CVF Conference on Computer Vision and
  Pattern Recognition}, pages 2818--2829, 2023.

\bibitem{chorowski2015attention}
Jan~K Chorowski, Dzmitry Bahdanau, Dmitriy Serdyuk, Kyunghyun Cho, and Yoshua
  Bengio.
\newblock Attention-based models for speech recognition.
\newblock {\em Advances in neural information processing systems}, 28, 2015.

\bibitem{cimpoi14describing}
M. Cimpoi, S. Maji, I. Kokkinos, S. Mohamed, , and A. Vedaldi.
\newblock Describing textures in the wild.
\newblock In {\em Proceedings of the {IEEE} Conf. on Computer Vision and
  Pattern Recognition ({CVPR})}, 2014.

\bibitem{cookjohn}
John~D Cook.
\newblock Upper and lower bounds for the normal distribution function.
\newblock {\em John D Cook's Blog}, 2018.

\bibitem{deshpande2021linearized}
Aditya Deshpande, Alessandro Achille, Avinash Ravichandran, Hao Li, Luca
  Zancato, Charless Fowlkes, Rahul Bhotika, Stefano Soatto, and Pietro Perona.
\newblock A linearized framework and a new benchmark for model selection for
  fine-tuning.
\newblock {\em arXiv preprint arXiv:2102.00084}, 2021.

\bibitem{devaguptapu2023deltapatching}
Chaitanya Devaguptapu, Samarth Sinha, K~J Joseph, Vineeth~N Balasubramanian,
  and Animesh Garg.
\newblock $\delta$-patching: A framework for rapid adaptation of pre-trained
  convolutional networks without base performance loss, 2023.

\bibitem{devlin2018bert}
Jacob Devlin, Ming{-}Wei Chang, Kenton Lee, and Kristina Toutanova.
\newblock {BERT:} pre-training of deep bidirectional transformers for language
  understanding.
\newblock In Jill Burstein, Christy Doran, and Thamar Solorio, editors, {\em
  Proceedings of the 2019 Conference of the North American Chapter of the
  Association for Computational Linguistics: Human Language Technologies,
  {NAACL-HLT} 2019, Minneapolis, MN, USA, June 2-7, 2019, Volume 1 (Long and
  Short Papers)}, pages 4171--4186. Association for Computational Linguistics,
  2019.

\bibitem{dosovitskiy2020image}
Alexey Dosovitskiy, Lucas Beyer, Alexander Kolesnikov, Dirk Weissenborn,
  Xiaohua Zhai, Thomas Unterthiner, Mostafa Dehghani, Matthias Minderer, Georg
  Heigold, Sylvain Gelly, Jakob Uszkoreit, and Neil Houlsby.
\newblock An image is worth 16x16 words: Transformers for image recognition at
  scale.
\newblock In {\em International Conference on Learning Representations}, 2021.

\bibitem{finn2017model}
Chelsea Finn, Pieter Abbeel, and Sergey Levine.
\newblock Model-agnostic meta-learning for fast adaptation of deep networks.
\newblock In {\em International conference on machine learning}, pages
  1126--1135. PMLR, 2017.

\bibitem{ghiasi2019fpn}
Golnaz Ghiasi, Tsung-Yi Lin, and Quoc~V Le.
\newblock Nas-fpn: Learning scalable feature pyramid architecture for object
  detection.
\newblock In {\em Proceedings of the IEEE/CVF conference on computer vision and
  pattern recognition}, pages 7036--7045, 2019.

\bibitem{gontijo2021no}
Raphael Gontijo-Lopes, Yann Dauphin, and Ekin~Dogus Cubuk.
\newblock No one representation to rule them all: Overlapping features of
  training methods.
\newblock In {\em International Conference on Learning Representations}, 2022.

\bibitem{guo2019spottune}
Yunhui Guo, Honghui Shi, Abhishek Kumar, Kristen Grauman, Tajana Rosing, and
  Rogerio Feris.
\newblock Spottune: transfer learning through adaptive fine-tuning.
\newblock In {\em Proceedings of the IEEE/CVF conference on computer vision and
  pattern recognition}, pages 4805--4814, 2019.

\bibitem{hariharan2015hypercolumns}
Bharath Hariharan, Pablo Arbel{\'a}ez, Ross Girshick, and Jitendra Malik.
\newblock Hypercolumns for object segmentation and fine-grained localization.
\newblock In {\em Proceedings of the IEEE conference on computer vision and
  pattern recognition}, pages 447--456, 2015.

\bibitem{he2021towards}
Junxian He, Chunting Zhou, Xuezhe Ma, Taylor Berg-Kirkpatrick, and Graham
  Neubig.
\newblock Towards a unified view of parameter-efficient transfer learning.
\newblock In {\em International Conference on Learning Representations}, 2022.

\bibitem{he2016resnet}
Kaiming He, Xiangyu Zhang, Shaoqing Ren, and Jian Sun.
\newblock Deep residual learning for image recognition.
\newblock In {\em 2016 {IEEE} Conference on Computer Vision and Pattern
  Recognition, {CVPR} 2016, Las Vegas, NV, USA, June 27-30, 2016}, pages
  770--778. {IEEE} Computer Society, 2016.

\bibitem{helber2019eurosat}
Patrick Helber, Benjamin Bischke, Andreas Dengel, and Damian Borth.
\newblock Eurosat: A novel dataset and deep learning benchmark for land use and
  land cover classification.
\newblock {\em IEEE Journal of Selected Topics in Applied Earth Observations
  and Remote Sensing}, 2019.

\bibitem{houlsby2019parameter}
Neil Houlsby, Andrei Giurgiu, Stanislaw Jastrzebski, Bruna Morrone, Quentin
  De~Laroussilhe, Andrea Gesmundo, Mona Attariyan, and Sylvain Gelly.
\newblock Parameter-efficient transfer learning for nlp.
\newblock In {\em International Conference on Machine Learning}, pages
  2790--2799. PMLR, 2019.

\bibitem{hu2021lora}
Edward~J Hu, yelong shen, Phillip Wallis, Zeyuan Allen-Zhu, Yuanzhi Li, Shean
  Wang, Lu Wang, and Weizhu Chen.
\newblock Lo{RA}: Low-rank adaptation of large language models.
\newblock In {\em International Conference on Learning Representations}, 2022.

\bibitem{iglovikov2018ternausnet}
Vladimir Iglovikov and Alexey Shvets.
\newblock Ternausnet: U-net with vgg11 encoder pre-trained on imagenet for
  image segmentation.
\newblock {\em arXiv preprint arXiv:1801.05746}, 2018.

\bibitem{ilharco2022patching}
Gabriel Ilharco, Mitchell Wortsman, Samir~Yitzhak Gadre, Shuran Song, Hannaneh
  Hajishirzi, Simon Kornblith, Ali Farhadi, and Ludwig Schmidt.
\newblock Patching open-vocabulary models by interpolating weights, 2022.

\bibitem{jaegle2021perceiver}
Andrew Jaegle, Felix Gimeno, Andy Brock, Oriol Vinyals, Andrew Zisserman, and
  Joao Carreira.
\newblock Perceiver: General perception with iterative attention.
\newblock In {\em International conference on machine learning}, pages
  4651--4664. PMLR, 2021.

\bibitem{jia2021scaling}
Chao Jia, Yinfei Yang, Ye Xia, Yi-Ting Chen, Zarana Parekh, Hieu Pham, Quoc Le,
  Yun-Hsuan Sung, Zhen Li, and Tom Duerig.
\newblock Scaling up visual and vision-language representation learning with
  noisy text supervision.
\newblock In {\em International Conference on Machine Learning}, pages
  4904--4916. PMLR, 2021.

\bibitem{jia2022visual}
Menglin Jia, Luming Tang, Bor-Chun Chen, Claire Cardie, Serge Belongie, Bharath
  Hariharan, and Ser-Nam Lim.
\newblock Visual prompt tuning.
\newblock In {\em European Conference on Computer Vision (ECCV)}, 2022.

\bibitem{KhoslaYaoJayadevaprakashFeiFei_FGVC2011}
Aditya Khosla, Nityananda Jayadevaprakash, Bangpeng Yao, and Li Fei-Fei.
\newblock Novel dataset for fine-grained image categorization.
\newblock In {\em First Workshop on Fine-Grained Visual Categorization, IEEE
  Conference on Computer Vision and Pattern Recognition}, Colorado Springs, CO,
  June 2011.

\bibitem{kirkpatrick2017overcoming}
James Kirkpatrick, Razvan Pascanu, Neil Rabinowitz, Joel Veness, Guillaume
  Desjardins, Andrei~A Rusu, Kieran Milan, John Quan, Tiago Ramalho, Agnieszka
  Grabska-Barwinska, et~al.
\newblock Overcoming catastrophic forgetting in neural networks.
\newblock {\em Proceedings of the national academy of sciences},
  114(13):3521--3526, 2017.

\bibitem{kolesnikov2020big}
Alexander Kolesnikov, Lucas Beyer, Xiaohua Zhai, Joan Puigcerver, Jessica Yung,
  Sylvain Gelly, and Neil Houlsby.
\newblock Big transfer (bit): General visual representation learning.
\newblock In {\em European conference on computer vision}, pages 491--507.
  Springer, 2020.

\bibitem{kornblith2019better}
Simon Kornblith, Jonathon Shlens, and Quoc~V Le.
\newblock Do better imagenet models transfer better?
\newblock In {\em Proceedings of the IEEE/CVF conference on computer vision and
  pattern recognition}, pages 2661--2671, 2019.

\bibitem{NEURIPS2020_bc573864}
Zhi Kou, Kaichao You, Mingsheng Long, and Jianmin Wang.
\newblock Stochastic normalization.
\newblock In H. Larochelle, M. Ranzato, R. Hadsell, M.F. Balcan, and H. Lin,
  editors, {\em Advances in Neural Information Processing Systems}, volume~33,
  pages 16304--16314. Curran Associates, Inc., 2020.

\bibitem{KrauseStarkDengFei-Fei_3DRR2013}
Jonathan Krause, Michael Stark, Jia Deng, and Li Fei-Fei.
\newblock 3d object representations for fine-grained categorization.
\newblock In {\em 4th International IEEE Workshop on 3D Representation and
  Recognition (3dRR-13)}, Sydney, Australia, 2013.

\bibitem{lee2019set}
Juho Lee, Yoonho Lee, Jungtaek Kim, Adam Kosiorek, Seungjin Choi, and Yee~Whye
  Teh.
\newblock Set transformer: A framework for attention-based
  permutation-invariant neural networks.
\newblock In {\em International conference on machine learning}, pages
  3744--3753. PMLR, 2019.

\bibitem{lee2019meta}
Kwonjoon Lee, Subhransu Maji, Avinash Ravichandran, and Stefano Soatto.
\newblock Meta-learning with differentiable convex optimization.
\newblock In {\em Proceedings of the IEEE/CVF conference on computer vision and
  pattern recognition}, pages 10657--10665, 2019.

\bibitem{lester2021power}
Brian Lester, Rami Al-Rfou, and Noah Constant.
\newblock The power of scale for parameter-efficient prompt tuning.
\newblock In {\em Proceedings of the 2021 Conference on Empirical Methods in
  Natural Language Processing}, pages 3045--3059, Online and Punta Cana,
  Dominican Republic, Nov. 2021. Association for Computational Linguistics.

\bibitem{li2021align}
Junnan Li, Ramprasaath Selvaraju, Akhilesh Gotmare, Shafiq Joty, Caiming Xiong,
  and Steven Chu~Hong Hoi.
\newblock Align before fuse: Vision and language representation learning with
  momentum distillation.
\newblock {\em Advances in neural information processing systems},
  34:9694--9705, 2021.

\bibitem{li2016revisiting}
Yanghao Li, Naiyan Wang, Jianping Shi, Jiaying Liu, and Xiaodi Hou.
\newblock Revisiting batch normalization for practical domain adaptation.
\newblock In {\em 5th International Conference on Learning Representations,
  {ICLR} 2017, Toulon, France, April 24-26, 2017, Workshop Track Proceedings}.
  OpenReview.net, 2017.

\bibitem{li2017learning}
Zhizhong Li and Derek Hoiem.
\newblock Learning without forgetting.
\newblock {\em IEEE transactions on pattern analysis and machine intelligence},
  40(12):2935--2947, 2017.

\bibitem{lin2017feature}
Tsung-Yi Lin, Piotr Doll{\'a}r, Ross Girshick, Kaiming He, Bharath Hariharan,
  and Serge Belongie.
\newblock Feature pyramid networks for object detection.
\newblock In {\em Proceedings of the IEEE conference on computer vision and
  pattern recognition}, pages 2117--2125, 2017.

\bibitem{liu2021swin}
Ze Liu, Yutong Lin, Yue Cao, Han Hu, Yixuan Wei, Zheng Zhang, Stephen Lin, and
  Baining Guo.
\newblock Swin transformer: Hierarchical vision transformer using shifted
  windows.
\newblock In {\em Proceedings of the IEEE/CVF International Conference on
  Computer Vision}, pages 10012--10022, 2021.

\bibitem{liu2022convnext}
Zhuang Liu, Hanzi Mao, Chao{-}Yuan Wu, Christoph Feichtenhofer, Trevor Darrell,
  and Saining Xie.
\newblock A convnet for the 2020s.
\newblock In {\em {IEEE/CVF} Conference on Computer Vision and Pattern
  Recognition, {CVPR} 2022, New Orleans, LA, USA, June 18-24, 2022}, pages
  11966--11976, 2022.

\bibitem{loshchilov2017decoupled}
Ilya Loshchilov and Frank Hutter.
\newblock Decoupled weight decay regularization.
\newblock {\em arXiv preprint arXiv:1711.05101}, 2017.

\bibitem{loshchilov2016sgdr}
Ilya Loshchilov and Frank Hutter.
\newblock {SGDR}: Stochastic gradient descent with warm restarts.
\newblock In {\em International Conference on Learning Representations}, 2017.

\bibitem{loshchcilov2019adamw}
Ilya Loshchilov and Frank Hutter.
\newblock Decoupled weight decay regularization.
\newblock In {\em International Conference on Learning Representations}, 2019.

\bibitem{mahajan2018exploring}
Dhruv Mahajan, Ross Girshick, Vignesh Ramanathan, Kaiming He, Manohar Paluri,
  Yixuan Li, Ashwin Bharambe, and Laurens Van Der~Maaten.
\newblock Exploring the limits of weakly supervised pretraining.
\newblock In {\em Proceedings of the European conference on computer vision
  (ECCV)}, pages 181--196, 2018.

\bibitem{maji2013fine}
Subhransu Maji, Esa Rahtu, Juho Kannala, Matthew Blaschko, and Andrea Vedaldi.
\newblock Fine-grained visual classification of aircraft.
\newblock {\em arXiv preprint arXiv:1306.5151}, 2013.

\bibitem{mallya2018piggyback}
Arun Mallya, Dillon Davis, and Svetlana Lazebnik.
\newblock Piggyback: Adapting a single network to multiple tasks by learning to
  mask weights.
\newblock In {\em Proceedings of the European Conference on Computer Vision
  (ECCV)}, pages 67--82, 2018.

\bibitem{Nilsback08}
Maria-Elena Nilsback and Andrew Zisserman.
\newblock Automated flower classification over a large number of classes.
\newblock In {\em Indian Conference on Computer Vision, Graphics and Image
  Processing}, Dec 2008.

\bibitem{herbarium}
New York Botanical~Garden (NYBG).
\newblock Fgvc 7 - herbarium 2020, 2020.

\bibitem{Pandy_2022_CVPR}
Michal P\'andy, Andrea Agostinelli, Jasper Uijlings, Vittorio Ferrari, and
  Thomas Mensink.
\newblock Transferability estimation using bhattacharyya class separability.
\newblock In {\em Proceedings of the IEEE/CVF Conference on Computer Vision and
  Pattern Recognition (CVPR)}, pages 9172--9182, June 2022.

\bibitem{parkhi12a}
Omkar~M. Parkhi, Andrea Vedaldi, Andrew Zisserman, and C.~V. Jawahar.
\newblock Cats and dogs.
\newblock In {\em IEEE Conference on Computer Vision and Pattern Recognition},
  2012.

\bibitem{quattoni2009recognizing}
Ariadna Quattoni and Antonio Torralba.
\newblock Recognizing indoor scenes.
\newblock In {\em 2009 IEEE conference on computer vision and pattern
  recognition}, pages 413--420. IEEE, 2009.

\bibitem{radford2021learning}
Alec Radford, Jong~Wook Kim, Chris Hallacy, Aditya Ramesh, Gabriel Goh,
  Sandhini Agarwal, Girish Sastry, Amanda Askell, Pamela Mishkin, Jack Clark,
  et~al.
\newblock Learning transferable visual models from natural language
  supervision.
\newblock In {\em International Conference on Machine Learning}, pages
  8748--8763. PMLR, 2021.

\bibitem{radford2018improving}
Alec Radford, Karthik Narasimhan, Tim Salimans, Ilya Sutskever, et~al.
\newblock Improving language understanding by generative pre-training.
\newblock {\em arXiv}, 2018.

\bibitem{rajbhandari2020zero}
Samyam Rajbhandari, Jeff Rasley, Olatunji Ruwase, and Yuxiong He.
\newblock Zero: Memory optimizations toward training trillion parameter models.
\newblock In {\em SC20: International Conference for High Performance
  Computing, Networking, Storage and Analysis}, pages 1--16. IEEE, 2020.

\bibitem{DBLP:journals/corr/SalehE15}
Babak Saleh and Ahmed Elgammal.
\newblock Large-scale classification of fine-art paintings: Learning the right
  metric on the right feature.
\newblock {\em International Journal for Digital Art History}, Oct. 2016.

\bibitem{sandler2022fine}
Mark Sandler, Andrey Zhmoginov, Max Vladymyrov, and Andrew Jackson.
\newblock Fine-tuning image transformers using learnable memory.
\newblock In {\em Proceedings of the IEEE/CVF Conference on Computer Vision and
  Pattern Recognition}, pages 12155--12164, 2022.

\bibitem{shen2021partial}
Zhiqiang Shen, Zechun Liu, Jie Qin, Marios Savvides, and Kwang-Ting Cheng.
\newblock Partial is better than all: Revisiting fine-tuning strategy for
  few-shot learning.
\newblock In {\em Proceedings of the AAAI Conference on Artificial
  Intelligence}, volume~35, pages 9594--9602, 2021.

\bibitem{touvron2022deit}
Hugo Touvron, Matthieu Cord, and Herv\'{e} J\'{e}gou.
\newblock Deit iii: Revenge of the vit.
\newblock In {\em Computer Vision – ECCV 2022: 17th European Conference, Tel
  Aviv, Israel, October 23–27, 2022, Proceedings, Part XXIV}, page 516–533,
  Berlin, Heidelberg, 2022. Springer-Verlag.

\bibitem{vasconcelos2022proper}
Cristina Vasconcelos, Vighnesh Birodkar, and Vincent Dumoulin.
\newblock Proper reuse of image classification features improves object
  detection.
\newblock In {\em Proceedings of the IEEE/CVF Conference on Computer Vision and
  Pattern Recognition}, pages 13628--13637, 2022.

\bibitem{vaswani2017attention}
Ashish Vaswani, Noam Shazeer, Niki Parmar, Jakob Uszkoreit, Llion Jones,
  Aidan~N Gomez, {\L}ukasz Kaiser, and Illia Polosukhin.
\newblock Attention is all you need.
\newblock {\em Advances in neural information processing systems}, 30, 2017.

\bibitem{WahCUB_200_2011}
C. Wah, S. Branson, P. Welinder, P. Perona, and S. Belongie.
\newblock Caltech-ucsd birds-200-2011.
\newblock Technical Report CNS-TR-2011-001, California Institute of Technology,
  2011.

\bibitem{wallingford2022task}
Matthew Wallingford, Hao Li, Alessandro Achille, Avinash Ravichandran, Charless
  Fowlkes, Rahul Bhotika, and Stefano Soatto.
\newblock Task adaptive parameter sharing for multi-task learning.
\newblock In {\em Proceedings of the IEEE/CVF Conference on Computer Vision and
  Pattern Recognition}, pages 7561--7570, 2022.

\bibitem{wang2019learning}
Haohan Wang, Songwei Ge, Zachary Lipton, and Eric~P Xing.
\newblock Learning robust global representations by penalizing local predictive
  power.
\newblock In {\em Advances in Neural Information Processing Systems}, pages
  10506--10518, 2019.

\bibitem{wang2022learning}
Zifeng Wang, Zizhao Zhang, Chen-Yu Lee, Han Zhang, Ruoxi Sun, Xiaoqi Ren,
  Guolong Su, Vincent Perot, Jennifer Dy, and Tomas Pfister.
\newblock Learning to prompt for continual learning.
\newblock In {\em Proceedings of the IEEE/CVF Conference on Computer Vision and
  Pattern Recognition}, pages 139--149, 2022.

\bibitem{xie2017aggregated}
Saining Xie, Ross Girshick, Piotr Doll{\'a}r, Zhuowen Tu, and Kaiming He.
\newblock Aggregated residual transformations for deep neural networks.
\newblock In {\em Proceedings of the IEEE conference on computer vision and
  pattern recognition}, pages 1492--1500, 2017.

\bibitem{yang2022rep}
Li Yang, Adnan~Siraj Rakin, and Deliang Fan.
\newblock Rep-net: Efficient on-device learning via feature reprogramming.
\newblock In {\em Proceedings of the IEEE/CVF Conference on Computer Vision and
  Pattern Recognition}, pages 12277--12286, 2022.

\bibitem{you2020co}
Kaichao You, Zhi Kou, Mingsheng Long, and Jianmin Wang.
\newblock Co-tuning for transfer learning.
\newblock {\em Advances in Neural Information Processing Systems},
  33:17236--17246, 2020.

\bibitem{yu2022coca}
Jiahui Yu, Zirui Wang, Vijay Vasudevan, Legg Yeung, Mojtaba Seyedhosseini, and
  Yonghui Wu.
\newblock Coca: Contrastive captioners are image-text foundation models.
\newblock {\em Transactions on Machine Learning Research}, 2022.

\bibitem{zeiler2014visualizing}
Matthew~D Zeiler and Rob Fergus.
\newblock Visualizing and understanding convolutional networks.
\newblock In {\em European conference on computer vision}, pages 818--833.
  Springer, 2014.

\bibitem{zhang2020side}
Jeffrey~O Zhang, Alexander Sax, Amir Zamir, Leonidas Guibas, and Jitendra
  Malik.
\newblock Side-tuning: a baseline for network adaptation via additive side
  networks.
\newblock In {\em Computer Vision--ECCV 2020: 16th European Conference,
  Glasgow, UK, August 23--28, 2020, Proceedings, Part III 16}, pages 698--714.
  Springer, 2020.

\bibitem{zhu2020deformable}
Xizhou Zhu, Weijie Su, Lewei Lu, Bin Li, Xiaogang Wang, and Jifeng Dai.
\newblock Deformable {\{}detr{\}}: Deformable transformers for end-to-end
  object detection.
\newblock In {\em International Conference on Learning Representations}, 2021.

\end{thebibliography}
}
\clearpage

\newcounter{appendixsection}
\renewcommand{\theappendixsection}{\Alph{appendixsection}}

\appendix
\setcounter{section}{0}
\renewcommand{\thesection}{\Alph{section}}
\renewcommand{\thesubsection}{\thesection.\arabic{subsection}}
\stepcounter{appendixsection}

{\LARGE\textbf{Appendix}}
\bigskip
\setcounter{page}{1}

Below we provide additional details and results which are not presented in the main manuscript.

\section[A]{Continual Learning with Open-\name }\label{sec:continual_learning}

With the Open-InCA adapter, each class prediction is isolated using a different dedicated query and classifier vector. For continual learning tasks, in addition to running multiple adapters in parallel as presented for multi-task results in Table \ref{tab:multitask}, Open-InCA enables an even more granular composition of adapter sub-tasks.   
Recall the Open-InCA adapter architecture is defined as 
\begin{align*}
&[v^1_\text{cross}(\mathbf{z}), \dots , v^c_\text{cross}(\mathbf{z})] := \operatorname{cross-attn}_\theta([z^1, \ldots, z^T], [q_1, \dots, q_c])\\
&\openinca_\theta(\mathbf{z}) := \operatorname{diag-head}_\theta \circ \operatorname{LN}   ([v^1_\text{cross}(\mathbf{z}), \dots, v^c_\text{cross}(\mathbf{z})])
\end{align*}
with $\operatorname{LN}$ denoting LayerNorm. Due to the properties of each operator, each class prediction can be computed separately as
\begin{align*}
\openinca &(\mathbf{z})_i = 
\langle W_i,\operatorname{LN}(\operatorname{cross-attn}_\theta([z^1, \ldots, z^T], [q_i])) \rangle.
\end{align*}
Because of this property we can remove a class prediction or add a new class prediction without any effects on other model predictions (as long as the parameters of $\operatorname{cross-attn}$ and $\operatorname{LN}$ remain fixed). As presented in Sec. \ref{sec:results}  we use ``query-only-training'' which trains new adapter classes while freezing $\operatorname{cross-attn}, \operatorname{LN}$ and enabling compatibility between task predictions.  

When training with ``query-only-training'' the softmax function, $\soft(u) = \dfrac{\exp(u^k)}{\sum_{i=1}^{c} \exp(u^i)}$, indirectly injects information of predictions from all classes due the normalization term in the denominator, which means gradients about a particular class $i$ will include information from other classes $j$. Instead, we can achieve complete training separation by using a Sigmoid final activation, $\sigma(u) = \dfrac{\exp(u)}{\exp(u) + 1}$, and a Binary Cross Entropy (BCE) loss that considers each prediction separately.
Clearly in ``query-only-training'' the adapter representation capacity is reduced, since the cross-attention weights are not trained. We present an experiment evaluating the performance of InCA, Open-InCA and ``query-only-training'' Open-InCA in Table \ref{tab:open_closed} and observe that despite the isolated and reduced parameter set in query-only-training of Open-InCA, the method is still competitive and outperforms Linear Probing on most datasets. 

Next we test Open-InCA for class-incremental learning, for which we consider the Split CIFAR-100 incremental learning benchmark. The Split CIFAR-100 dataset is trained with 10 incremental learning episodes each introducing 10 new classes. As in \cite{chaudhry2018efficient}, we present the average episode accuracy and forgetting of ``query-only-training'' Open-InCa and additional baselines.

In particular we evaluate Open-InCa using a ViT-B/16 along with state of the art methods L2P \cite{wang2022learning}, LwF \cite{li2017learning} and EWC \cite{he2021towards}. Nonetheless, we do not apply any special routing of our learned episodic models and simply combine their predictions. In contrast L2P is a prompt based approach that, during inference, passes each new sample to an auxiliary classifier to predict its corresponding episode (in this case a 10-way classifier) and the corresponding episode model is up-weighted according to the prediction. We believe that with such an auxiliary classifier Open-InCA performance can significantly improve, nonetheless we observe that Open-InCa can simply leverage a larger model efficiently to achieve state of the art accuracy.  We leave routing of samples to different learned sub-models as an interesting avenue for future work. \par

In addition, Open-InCA has additional benefits as compared to typical class-incremental learning approaches:
\begin{itemize}
        \item \textbf{Flexible incrementation} With Open-InCA different episodes can naturally contain a variable number of classes and episodes can be further decomposed if needed. This is since one can modify the model at the granularity of a single-class predictor via the Open-InCA adapter architecture by introducing or removing additional $q_i$ and $W_i$.
    \item \textbf{Reduced forgetting risk} With Open-InCA the ability of adding new classes without forgetting is built-in into the architecture, as prediction of different classes ensures that the previous class predictions remain the same (\ie, no logit regression) which reduces catastrophic forgetting. 
    \item \textbf{Parameter and computation efficient} The Open-InCA adapter benefits from the InCA approach, which is parameter efficient and computationally efficient during inference (see Table \ref{tab:multitask}) as well as during training (see Fig. \ref{fig:opt_speed} for comparison with prompts).
\end{itemize}

\begin{figure}[htbp]
\centering
\begin{minipage}{0.45\textwidth}
   \resizebox{1\textwidth}{!}{
    \begin{tabular}{ccc}
\toprule
              Method & \makecell{Average \\ Accuracy ($\uparrow$) } & Forgetting ($\downarrow$) \\
\midrule
      LP-sequential$^{*}$  &            17.7 &      59.1 \\
              Full-FT-sequential$^{*}$ &            33.6 &      86.9 \\
                EWC \cite{kirkpatrick2017overcoming}  &            47.0 &      33.3 \\
                  LwF \cite{li2017learning} &            60.7 &      27.8 \\
                L2P \cite{wang2022learning}  &            \bf83.8 &       \bf7.6 \\
                \midrule
 Open-InCA (ViT-B/16) &           \bf 83.0 &      \bf 9.1 \\
 Open-InCA (ViT-L/16) &            \bf 88.3 &       \bf 7.1 \\
Open-InCA  (ViT-H/14) &                \bf 86.1  &           \bf 8.2 \\
\bottomrule
\end{tabular}}
\captionof{table}{\textbf{CIFAR-100 Class-Incremental Learning} Split CIFAR-100 is trained with 10 episodes of 10 classes in the standard CIL evaluation suite \cite{wang2022learning}. Average accuracy and forgetting is reported over the 10 episodes according with \cite{chaudhry2018efficient}. $^{*}$Sequential fine-tuning results are taken from \cite{wang2022learning}.} 
\label{tab:my_label}
\end{minipage}%
\hspace{1cm}
\begin{minipage}{0.45\linewidth}
  \centering
    \setlength{\tabcolsep}{4pt}
   \resizebox{1\textwidth}{!}{
    \begin{tabular}{c| cccc}
\toprule
{}& \multicolumn{4}{c}{\textbf{Top-1 Test Error}}\\
       \midrule
       Dataset & InCA & Open-InCA & \makecell{Query only \\ Open-InCA} & In. LP \\
       \midrule
       CUB-200 &  \bf 9.1 &       9.5 &                   12.1 & 16.2 \\
           DTD & 17.8 &     \bf 17.1 &                 19.2 & 18.9 \\
     Aircrafts & \bf 15.8 &      18.1 &                 38.6 & 50.6 \\
        MIT-67 & 10.1 &        9.4 &                  \bf 9.1 & 9.7 \\
Oxford Flowers &  \bf 0.3 &       0.4 &                  0.4 & 0.6 \\
   Oxford Pets &  4.7 &         \bf 4.0 &                  5.4 &  6.1\\
   Stanf. Cars &  \bf 8.4 &      \bf 8.4 &                 22.8 & 29.2 \\
   Stanf. Dogs &  8.1 &       5.7 &                  \bf 5.3 & \bf 5.3 \\
   \midrule
      Average  &  9.3 &       \bf 9.1 &                  14.1 &  17.1 \\
\bottomrule
\end{tabular}
}
\captionof{table}{\textbf{Open-InCA adapter performance} We compare InCA, Open-InCA, ``query-training'' Open-InCA and Intermediate Linear Probing (In. LP). We observe that Open-InCA is comparable with InCA and that ``query-training'' significantly out-performs In. LP. }\label{tab:open_closed}
\end{minipage}
\end{figure}

\section{Intermediate Representation Signatures}\label{sec:intermediate_rep}
The parallel training of InCA results in the synthesis of tens of models that can run inference with insignificant per-adapter marginal costs. As a result we have the ability to glean highly useful information about the network's different representations and study the network inner representations effectively. This is especially important for recent non-convolutional based architectures that do not have as many inductive biases explaining some of their behavior. In this section we present results showing information we retrieve from the performance of InCA adapters.
\begin{figure*}[ht]
    \centering
    \includegraphics[width=10cm]{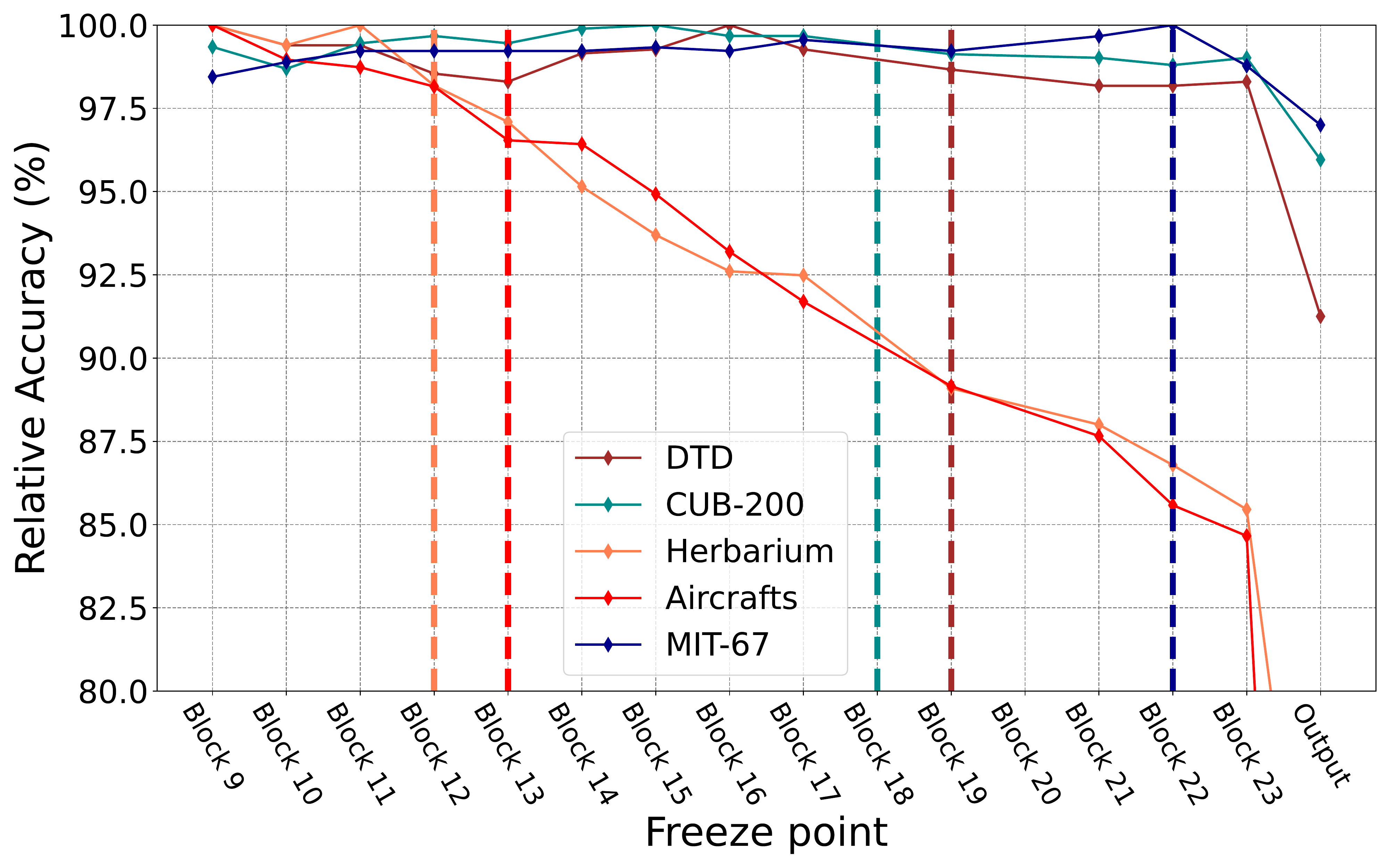} \\
    \caption*{Figure 4: (repeated) \textbf{Partial fine-tuning vs. InCA} Vertical dashed lines indicate the top InCA layer; curves show final test accuracy for different partial tuning training runs. Each mark indicates a run where all of the pre-trained model parameters are trained up to a ``freeze point'' in the network's layers.  Note partial tuning performance saturates in close proximity to the optimal InCA adapter layer. This is aligned with our hypothesis that full fine-tuning attempts to surface \emph{existing representations already in the network}. In that case, performance improves until the tuning approach unlocks the capacity to utilize an existing relevant representation and performance plateaus afterwards.}
\end{figure*}

\subsection{Partial fine-tuning and adapter performance}
Below we present in detail the experiments discussed in Sec. \ref{sec:analysis}, in particular regarding partial fine-tuning and \name.  The experiments illustrate the relationship between InCA adapters at different layers with partial tuning. We tune the pre-trained model starting from different ``freezing points''. In particular for neural network $f(x) = g_1 \circ \dots g_l$, for each freezing point $g_m$ we consider its position $m$ and all of the preceding layers and apply layer freezing to $g_1, \dots g_{m-1}$ (\ie~not updating gradients for those layers). Back-propagation is then only applied for optimizing $g_m, \dots g_l$ including the network's prediction ``head''.  In Figure \ref{fig:partial_tuning} we show the dynamics of partial tuning, where we optimize the pre-trained network (ViT-L/16) in different runs with each run having a different freezing point. We compile the final Top-1 Test accuracy of each freezing run to create a partial tuning ``curve'' for a single dataset. We compare the partial tuning performance curve of each dataset with the corresponding top layer of the InCA adapter trained on that dataset and observe that they are highly aligned, with datasets that prefer later InCA layers plateauing in their test accuracy earlier (at a later freezing point). In particular what we observe is that partial tuning performance plateaus roughly at the same layer where InCA identifies the top adapter representation. This is also the point at which partial fine-tuning is capable of harnessing that representation for the downstream task. 
Overall this gives further evidence that \emph{``your representations are in the network''} and fine-tuning simply surfaces existing representations that are already identified by InCA. When drawing the vertical lines of the top \name adapters, we refer to output layers, \eg,  the adapter at block 19 means the adapter corresponding to the final output of block 19, or the first input of block 20. 

\subsection{Task Layer Affinities}\label{subsec:inter}
In \name we select top-performing adapters that ``listen'' to different intermediate representations of a neural network. In our work we observe that one is able to achieve strong and diverse transfer learning by utilizing intermediate representations, and that for challenging tasks it is often required to use intermediate representations to achieve top results. Indeed the best representation layer for an adapter tends to be highly robust to hyper-parameter variables of the optimization. Even more intriguingly, we find that this representation affinity is preserved across different pre-trainings and even architectures. This is, certain tasks have a strong \emph{``affinity''} to a certain range of representation layers even for different architectural circumstances. The majority of the architectures we consider have some pre-trained component on one of the ImageNet datasets (aside from the CLIP ViT-L/14 model). At the same time, the fact that different architectures give rise to similarly helpful representations gives strong clues about the effect of different architectures as compared with the pre-training task during learning over a large diverse dataset such as ImageNet. \par

\begin{figure}[b]
    \centering
    \includegraphics[width=7cm]{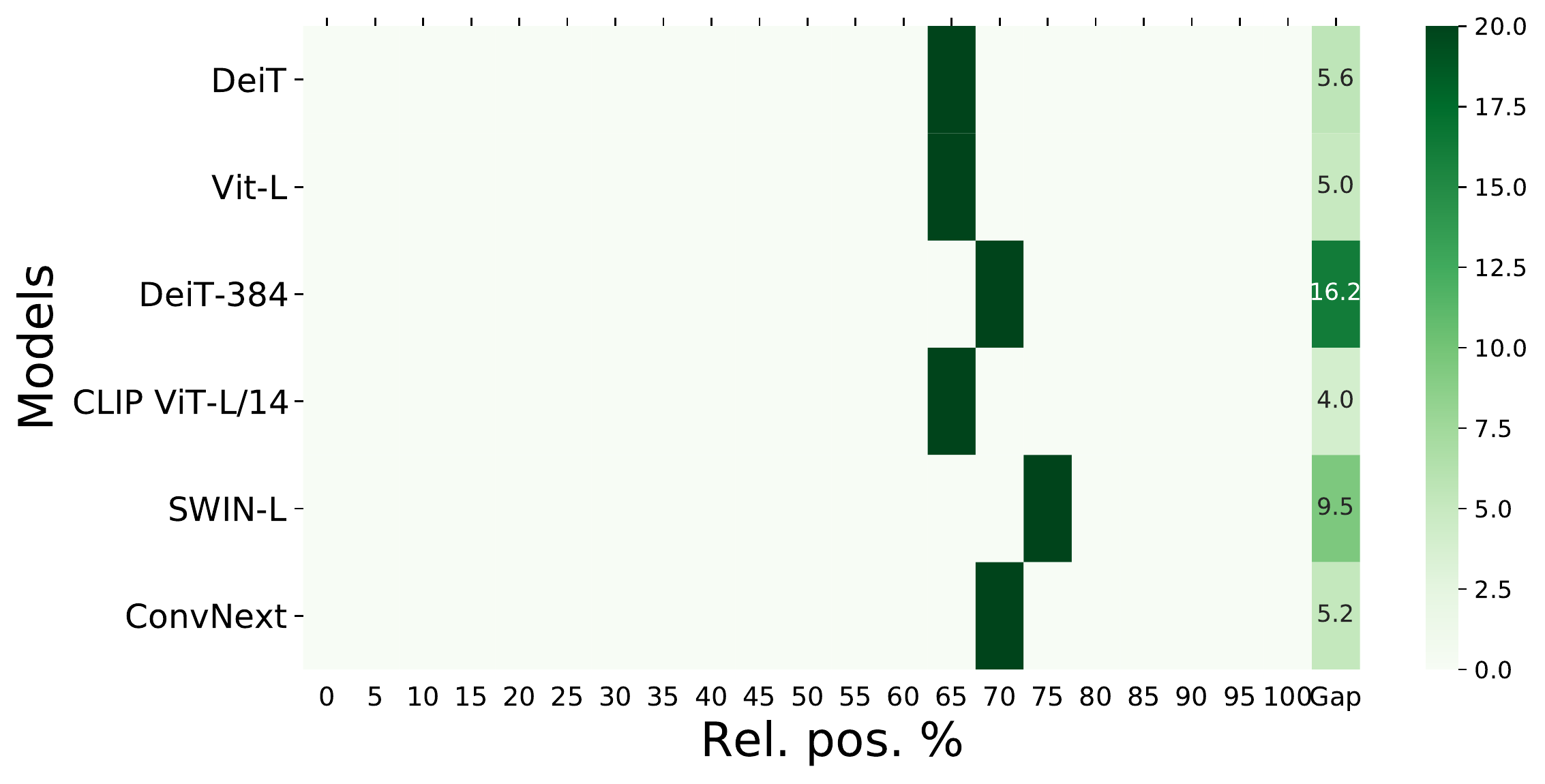}%
    \includegraphics[width=7cm]{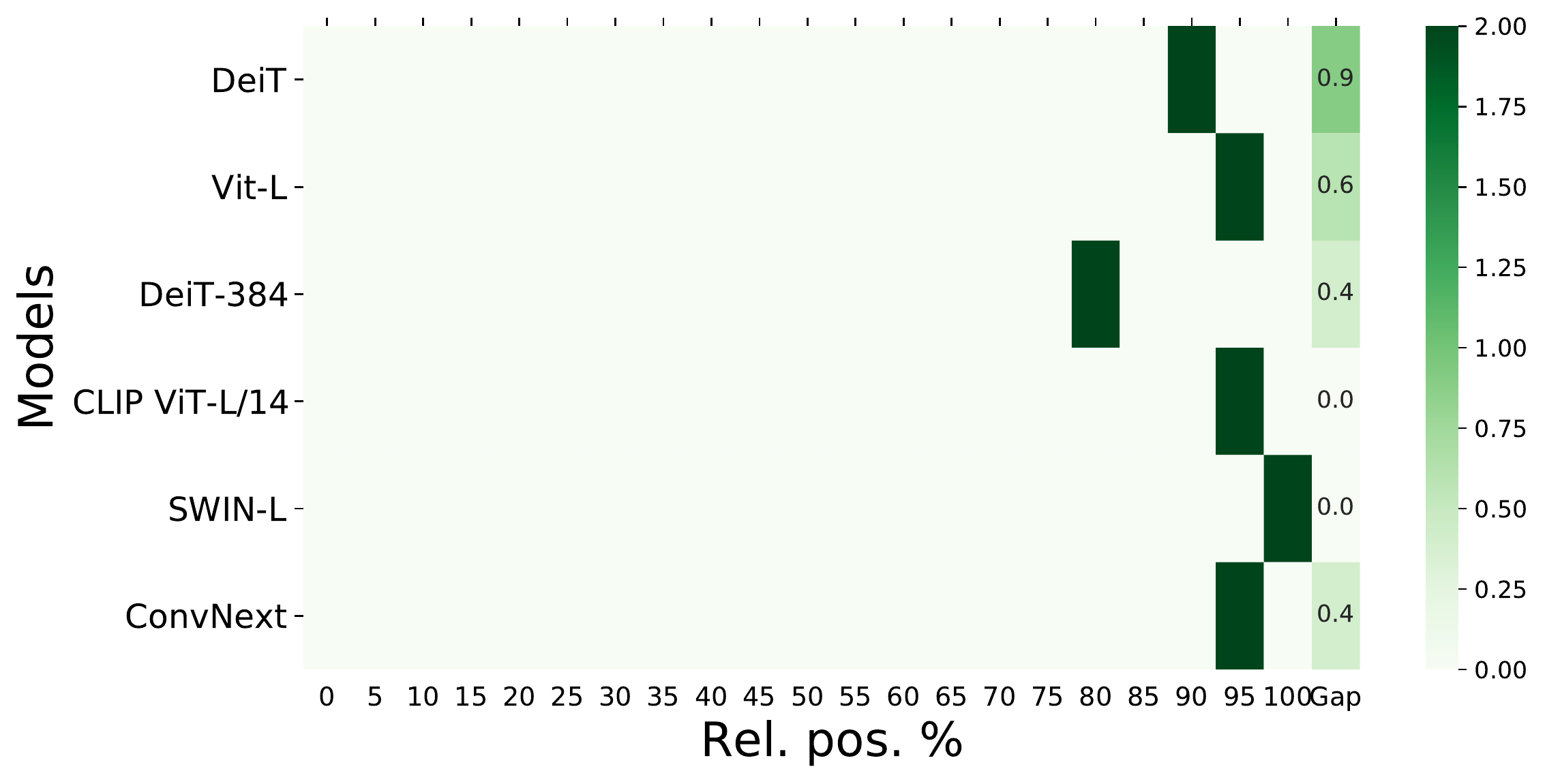} \\
    \includegraphics[width=7cm]{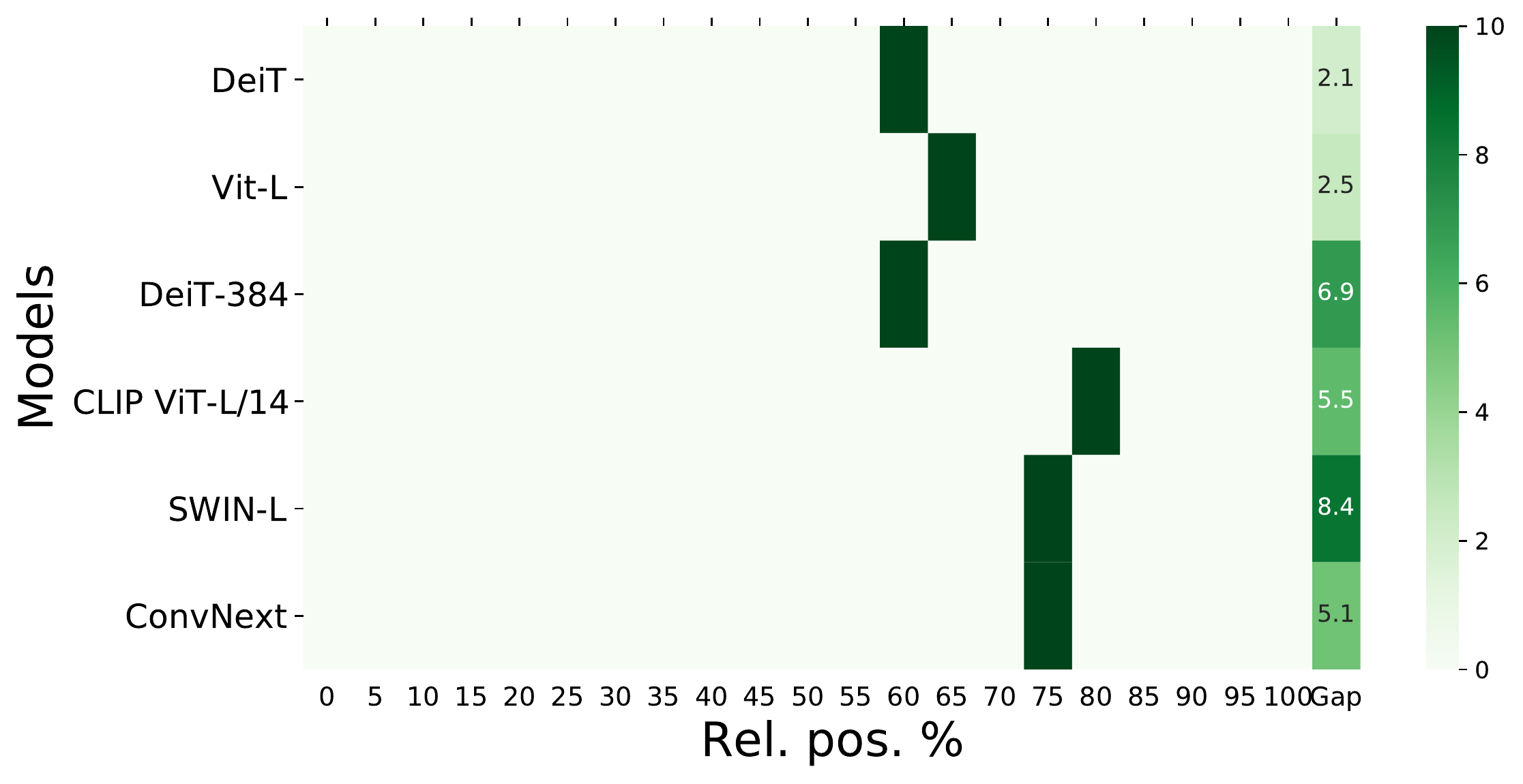} \\

    \caption{\textbf{Best-performing representation for InCA adapter for Aircraft (Top-left) MIT-67 (Top-right)
and Stanf. Cars (Bottom).}}
\label{fig:inca_layer_affinity}
\end{figure}

In detail, we look at the best-performing \name adapter for a fixed task on different architectures. The pre-trained models we consider consist of 2 different pre-trainings of the ViT-L/16 architecture (ViT-original and DeiT), the 384-resolution pre-training of DeiT with the resolution adjusted ViT-L/16, CLIP's ViT-L/14 architecture, the SWIN-L architecture, and the convolutional based ConvNext-Base architecture. All of the vanilla ViTs we consider each have 24 residual transformer blocks so that comparing between blocks is directly aligned. SWIN-L and ConvNext follow the ``Stage'' breakdown of blocks, namely SWIN-L has (2,2,18,2) stage breakdown that conveniently also adds up to 24 blocks (hence aligned in the figure) and lastly, ConvNext follows a (3-3-26-3) + head stage block composition, which we rescale in the figures to fit on the same 24 block range. In addition in the plots we also present the test error gap of each architecture with using the \name adapter applied on its final block representation. Tasks that prefer earlier layers such as Stanf. Cars and Aircraft have a large gap from the performance of the last layer representation adaptation and such later layers lead to sub-optimal results.

We remark that the work of \name sheds light on the inner representations learned by neural networks showing in some aspects performance is invariant to the architecture and more based on the pre-training dataset. We leave this topic for further research and find it to be an intriguing topic of study.

\section{Efficiency Results}\label{sec:eff_results}
InCA is highly efficient especially for large models, which is based on the isolated adapter architecture that does not modify the backbone. We delineate the efficiency aspects as follows:
\begin{itemize}
    \item \textbf{Training memory efficiency} The use of a frozen pre-trained model makes the training much more efficient and scalable since not all of the intermediate computations need to be stored as done in standard training or as required by methods that compute gradient information using inner-layers of the network. As soon as \emph{any} intermediate layer requires a gradient, \emph{all} subsequent activations must be held in GPU memory after the forward pass. This means methods like LoRA, FitBit and VPT all require storing of all of the activation maps for all of the layer operations in the network since they update parameters based on gradients from the very early layers in the network.
    
    \item \textbf{Fast optimization} Unlike typical parameter efficient methods that insert some form of trainable parameters in the network, InCA adapters are trained with ``direct gradient'' information coming from an isolated loss. Essentially each adapter corresponds to a very shallow neural network trained directly via back-propagation. This makes the training dynamics fast as direct gradient information about the loss easily reaches all of the adapter parameters. On the other hand, to update inserted parameters in the backbone, the gradient information is indirect and needs to be back-propagated through the backbone, with the risk of information loss and making the optimization more challenging, as we and the authors \cite{jia2022visual} observe regarding prompt tuning. 
    \item \textbf{Efficient multi-task inference} As we present in Table \ref{tab:multitask} the unchanged backbone execution enables efficient and parallel inference efficiency as multiple tasks can be evaluated at once.
\end{itemize}

\subsection{Computational Efficiency of \name Compared with VPT}\label{subsec:vpt_eff}
In Table \ref{tab:inca_vpt_eff} we observe that \name~is an order of magnitude more efficient to train than VPT . For the results in the table we consider the VPT-Deep adaptation method, that is trained with 50 prompt tokens in each layer. We report calculated training times in GPU-hours of a standard Nvidia-T4 GPU using a ViT-L/16 architecture and accuracy numbers based on the datasets of Table \ref{tab:fine_grained_vit} with the DeiT pre-training. For larger architectures such as ViT-H/14 (``ViT Huge'') the difference in training-time is even more striking, as \name maintains good per-run training time of 2.5, VPT-Deep requires staggering 55.8 GPU-hours per-run for a single GPU. On ViT-H/14 this is exacerbated as we must reduce the batch-size of VPT significantly to fit training on a common-place single GPU (Nvidia-T4). We measure in terms of training InCA and VPT-Deep for the same number of epochs. This however, is inaccurate as InCA trains an order of magnitude faster on a per epoch basis (see Figure \ref{fig:opt_speed}).

\begin{table}[H]
    \centering
   \resizebox{0.7\textwidth}{!}{
   \setlength{\tabcolsep}{5pt}
    \begin{tabular}{ccccccc}
\toprule
           Method & \makecell{Mean \\ Test Err.} & \makecell{Max. \\
           Full-FT \\ gap} & \makecell{Training time \\ per run \\ (GPU hrs.)} & \makecell{\# Hparam.  \\ per dataset} & \makecell{Train time \\ per dataset \\ (GPU hrs.)} \\
\midrule
             InCA &           \bf10.2 &                       \bf2.4 &                          \bf2.0 &                 \textbf{2} (parallel) &                            \bf{4.0 (2.4$^*$)} \\
VPT Deep  \cite{jia2022visual}  &           12.3 &                       6.8 &                       5.8 &                           24 &                               139.6 \\
\bottomrule
\end{tabular}
}\vspace{0.2cm}
 \caption{\textbf{Computation costs of adaptation} We adapt ViT-L/16 to CUB-200 downstream classification with the same number of training epochs. We evaluate the training and computational costs of a single run and training VPT-Deep and \name for one training dataset. $^*$Training with 2 learning rates in parallel leads to training time decreasing from 4.0 to 2.4 GPU-hours.}
    \label{tab:inca_vpt_eff}
\end{table}

We attribute the difference in training time of \name to:
\begin{enumerate}
    \item \name does not require back-propagation through the whole model which gives $\sim 50\%$ speed improvement alone.
    \item \name is robust to hyper-parameters and we optimize it using just 2 learning rates, compared with the hyper-parameter set of VPT (in our experiments we use 24 hyper-parameter configurations per dataset while using the full configuration presented in \cite{jia2021scaling} takes even longer). In addition, with ``one-to-many'' training, we train the two hyperparameters of InCA in parallel and report the specific training time in Table \ref{tab:inca_vpt_eff} denoted by ($^*$) for parallel hyper-parameter training.
    \item \name does not increase the number of propagated tokens in the transformer (\eg~in VPT with 100 propagated tokens the attention matrix doubles, from $\sim 4\times 10^4$ to $\sim 9 \times 10^4$ entries).
\end{enumerate}

\subsection{Optimization dynamics of \name and VPT}\label{subsec:opt_dynamics}
\begin{figure}[ht]
    \centering
    \includegraphics[width=8cm]{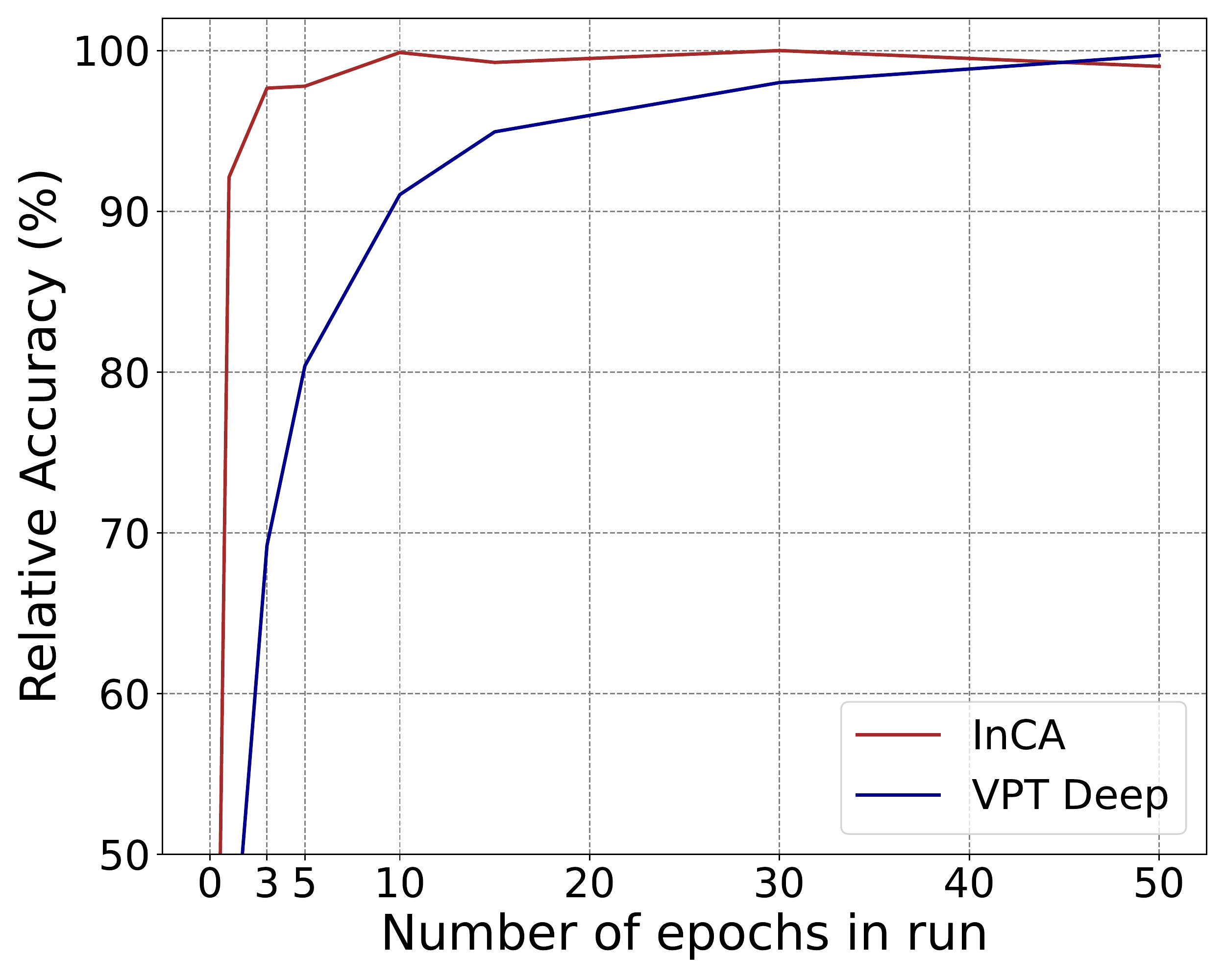} \\
   
    \caption{\textbf{Optimization Speed} for training InCA and prompt tuning (VPT-Deep) on the Aircrafts dataset. We train each method until completion with varying numbers of epochs and report the relative final test accuracy to 50 epoch training. The shallow adapter architecture and direct gradient signal in InCA makes the training of the adapter an order of magnitude faster (in terms of gradient updates) than prompt tuning approaches. Both methods use batch-size 32 and take the same number of gradient steps in each corresponding run, under the optimal learning rate.}
\label{fig:opt_speed}
\end{figure}

In \cref{fig:opt_speed} we conduct an experiment where we train \name and state of the art prompting method VPT-Deep \cite{jia2022visual} for different numbers of epochs and report the final test accuracy. We observe that InCA trains order of magnitude faster than prompting and reaches within $95\%$ relative test accuracy after 3 training epochs.  

\section{Theoretical Analysis}\label{sec:theory}
Empirically, we consistently observe that cross-attention as opposed to a linear or MLP-3 architecture enables \name~to better harness the existing model. We present a theoretical result asserting that using the cross-attention layer for aggregation as opposed to linear averaging, or even full-concatenation followed by a large dimensional linear layer is capable of learning over a strictly broader set of data distributions. 

We give the precise statement in Theorem \ref{thm:cross_advantage} and intuitively argue that cross-attention with learned queries has the ability to sift through irrelevant pieces of the representation that may be at \emph{variable} positions in different data samples.

Recall in the settings considered thus far, the extracted activation of an image data-point can be viewed as $\xb_i \in \RR^{d \times T}$ or $T$ tokens \eg~$\xb_i = [x_i^1, x_i^2 \dots  x_i^T]$, with $x_i^j \in \RR^{d}$. We argue that in many scenarios, task-pertinent information is a property of individual \emph{tokens} (\eg~ $x_i^j$) within a data-point $\xb_i$ and not a property of the overall feature map. We present the theorem below.
To this end we define a Token-Separability (TS) notion of a dataset.

\begin{definition}[Token-separable Dataset]\label{def:ts_dataset}
A dataset $\calD = \{(\xb_1, y_1), \dots (\xb_n, y_n) \}$ with $\xb_i = [x_i^1, x_i^2, \dots x_i^T] \in \RR^{d \times T}$ and $y_i \in \{-1,1\}$ is said to be linearly-token-separable if there exists a scalar $c > 0$ and $w \in \RR^d$ satisfying $\|w\|_2=1$, such that for each data point $(\xb_i, y_i) \in \calD$ there exists a token $x_i^{j_i} \in \xb_i$ with
\begin{equation}
    y_i (\langle x_i^{j_i}, w \rangle + b) \ge c.
\end{equation}
We define $(w_\calD, b_\calD)$ and $c_0$ as the maximum margin solution and maximum margin respectively, i.e. $c_0 = \max_{\{\|w\| = 1, b\in \RR \}} \min_{\calD} \{y_i (\langle x_i^{j_i}, w \rangle + b) \}$ ~for $\calD$ with $(w_\calD, b_\calD)$ corresponding to the selected $c_0$.
\end{definition}
Intuitively, $\calD$ is a TS-dataset if each of its data points contain a token that leads to linear separability (the same $w$ shared among all points $\xb_i \in \calD$ ).
One can further distinguish between \emph{aligned}-TS Datasets, where the index $j_i$ of the linearly separating token is consistent among the $n$ data points, or \emph{permutable}-TS where $j$ is dependent on $i$. Further TS datasets can be generalized to $k$-token separable datasets where $k$ tokens are responsible for separability in each $\xb_i$, for this theoretical contribution we don't make an assumption on whether the dataset is aligned or permuted, but consider the setting provided by Definition \ref{def:ts_dataset} (i.e. not the $k$-separable generalization).
We present an analytical statement for the advantage of $\ca$, the theorem is provided for binary classification via a scalar prediction, but can conventionally extend to $C$-class classification. For binary classification we define a prediction via the standard scalar binary aggregator as $\sigma(u) = \mysign (\sum_i u_i)$ that converts a vector into a binary prediction.

\begin{restatable}{theo}{main}
Let $\calD$ be a binary-class, token-separable dataset with max-margin $c_\calD$ and max-margin solution $(w_\calD, b_\calD)$ consisting of $n$ data points.  Suppose that $\calD$ is distributed such that for $\xb_i=(x_i^1, \dots x_i^T)$ with $(\xb_i, y_i) \in \calD$ are normalized for separating token $x_i^{j_i} \in \calD$ and that the rest of the tokens correspond to ``noise'', $x_i^k \sim N\big(0, I / d\big)$, ~$\EE \|x_i^k\|^2_2 = 1$. 
Furthermore assume
\begin{equation}
  c_{\calD} \ge \max\bigg(\sqrt{ \frac{32}{d} \big( \log(1/\delta) + \log(2nT)\big)},~ 2 |b_{\calD}| \bigg).
\end{equation}
Then there exists a cross-attention classifier
\begin{equation}
    f(x; q, \{W\}, b) = \sigma\parens{\sum_{l = 1}^T \ca(\xb, q)_l + b}
\end{equation}
that separates $\calD$ with probability at least $1-\delta$. 
In contrast, every fixed member $g(\xb; w, b) = \sigma\parens{\sum_{l = 1}^T \xb_i \cdot w + b}$ of the linear classifier family will fail to separate $\calD$ with probability at least 
$\frac{1}{\sqrt{2\pi}} \frac{s}{s^2 +1} \exp(-s^2/2)$
where $s = \frac{\sqrt{d}}{\sqrt{T - 1}} c_\calD$.
\label{thm:cross_advantage}
\end{restatable}

As stated above, the failure probability of the simple linear classifier $g$ depends on $s$ which satisfies $s \sim \sqrt{d/ T}$. For existing architectures $d, T$ tend to have a similar order of magnitudes, e.g. for ViT-B/16, $d=768, T=196$ which makes the failure probability non-negligible.
Before presenting the proof, we make the following observation: in \name, we use the same $\ca$ layer with latent $q$, which we show can be simplified via reparameterization.
\begin{obs}[Query Collapse Reparameterization]\label{thm:query_colapse}
A single-head cross-attention parameterization with latent $[q]$ is equivalent to the following simplified layer,
$\ca(\xb, q) = \sum \soft(q^*  \xb) \odot \mathbf{W}\xb$ with $q^* \in \mathbb{R}^{d}$.
\end{obs}
This can be derived by decomposing the attention score which is the input to $\soft$. 
\begin{align*}
    a_{j} = \langle \Wb_q q ,\Wb_k x^j \rangle = (\Wb_q  q)^\top (\Wb_k x^j) = \\
     q^\top \Wb_q^\top \Wb_k x^j =  (q^{*})^{\top} x^j .
\end{align*}
Where $q^* = q^\top \Wb_q^\top \Wb_k$ and $q^* \in \RR^{d}$, hence the $\ca$ layer simplifies, which is used in the proof.

As our proof shows, the $\ca$ layer can operate on a large data bandwidth, \eg $\xb \in \RR^{d \times T}$ while still being selective in finding task specific representations. 
Empirically we also observe that increasing the number of heads of $\ca$ also improves the performance of the \name~. 
This is in part because it enables the learned latent query parameter $q$ to identify more useful token patterns, and since $q$ is fixed using more heads remain stable (as opposed to when $q$ is a data input).
We now present the proof of the claim.

\paragraph{Proof of Theorem \ref{thm:cross_advantage}} \label{sec:proof}
\begin{proof}
The proof of the theorem has two parts A) the positive condition on the $\ca$ layer and B) the negative condition on the linear layer (a non-separability probability lower bound). We start with A) and consider separability of positive and negative data samples in turn. First we simplify and write an equivalent cross-attention binary classifier expression for the $\ca$ classifier.

\paragraph{Positive result for the cross-attention model}
We consider a ``single-head'' cross-attention layer and by Observation \ref{thm:query_colapse} we can write the $\ca$ layer as follows
\begin{equation}
    \ca(\xb_i; q, \{\Wb\}) = \sum_{j=1}^{T} \soft (\langle x_i^j,  q^* \rangle ) \cdot \Wb_v x_i^j.
\end{equation}
Note that $\soft$ is a function of the entire vector $\{\langle x_i^j,  q^* \rangle\}_{j \in [1,T]}$, however we write it in the form above to illustrate the summed terms.
For simplicity of notation, we drop the asterisk and write $q^* \in \RR^{d}$ as $q$. Combining $\ca$ with the binary aggregator, we have aggregation over the output vector of the $\ca$ layer.
\begin{gather*}
    f(\xb_i ; q, \{W\}, b) \\
    = \sigma \bigg(\sum_{l=1}^{d} \big(\ca(\xb_i ; q, \{W\}) \big)_l + b \bigg).
\end{gather*}
Define $S(\xb_i, q) \in \mathbb{R}^{1 \times T}$ as the computed $\soft$ argument,
\begin{equation}
    S(\xb_i, q) = S = \soft ( [\langle x^1_i, q \rangle,  \dots \langle x^T_i,  q \rangle ] ).
\end{equation}
Substituting into the classifier, we have
\begin{align*}
        f(\xb_i; q, \{\Wb\}, b) &=\sigma \bigg(\sum_{l=1}^{d} \big( \sum_{j=1}^{T} S_j \cdot  \Wb_v x_i^j  \big)_l + b\bigg) \\
        &=\sigma \bigg( \sum_{j=1}^{T}  S_j \sum_{l=1}^{d} (\Wb_v x_i^j )_l + b \bigg).
\end{align*}
Let $u = \sum_{l=1}^{d} (\Wb_v)_{[l,:]}$ be the sum of the rows of $\Wb_v$. Note  $x_i^j$ can be pulled out from the inner summation to give
\begin{align*}
            f(\xb; q, u, b) =\sigma \bigg( \sum_{j=1}^{T}  S_j \langle u, x_i^j \rangle + b \bigg).
\end{align*}
Thus the $\ca$ classifier presented is equivalent to the parameterization above.
Next we consider the two terms in the sum, namely $S_j$ and $\langle u, x^j_i \rangle$. We will be deriving their distribution in the case where a data point $(\xb_i, y_i)$, has prediction labels $y_i = 1$ and $y_i = - 1$ separately.
We start with $y_i = 1$ and consider
\begin{equation}
    S_k = \frac{\exp(\langle x_i^k, q \rangle)}{\sum_{j=1}^{T} \exp (\langle x^j_i, q \rangle)}.
\end{equation}
Take $j_i$ to be the separating token for sample $\xb_i$.  By the assumption of the theorem for $k \ne j_i$ the tokens correspond to isotropic noise of expected squared norm 1, i.e. $\xb_i^k \sim N(0, I/d)$.
For a fixed $u \in \mathbb{R}^{d}$ with $\|u\|_2 = 1$, we take $\eta_k$ to be the distribution of the dot product,
\begin{equation}
   \eta_k =  \langle u, \xb_i^k \rangle = \sum_l  (u)_l \cdot (\xb_i^k)_l.
\end{equation}
For $k \ne j_i$ this is a sum of independent Gaussians and each coordinate is distributed as $\sim N(0, \frac{u_l^2}{d})$. As such we have 
\begin{align*}
    \langle u, \xb_i^k \rangle  \sim N\parens{0, \sum_l \frac{1}{d} \cdot u_l^2} &= N(0, \|u\|_2^2/d) \\
    &= N\parens{0, 1/d}
\end{align*}
since $\|u\|_2 = 1$.
Next we consider $k = j_i$.  By the hypothesis we have that
\begin{equation*}
    y_i (\langle \xb_i^{j_i}, w_\calD \rangle + b_\calD) \ge c_{\calD}.
\end{equation*}
With positive label ($y_i = 1$) this gives $\langle \xb_i^{j_i}, w_\calD \rangle + b_\calD \ge c_{\calD}$.  Note that since $c_\calD \geq 2 |b_\calD|$ we have that $\langle \xb_i^{j_i}, w_\calD \rangle \geq c_\calD / 2$.  Since $w_\calD$ is the maximal margin solution, we have $\|w_\calD\| = 1$ and $c_\calD >0$.
Take $q$ to be of the form $q = t \cdot  w_\calD$ for $ t \in \RR^{+}$ and $u = w_\calD$, then
\begin{equation}
    t \cdot \eta_{j_i} = \langle x_i^{j_i}, q \rangle = t \langle x_i^{j_i}, w_\calD \rangle \ge t \cdot c_{\calD} / 2.
\end{equation}
For $\eta_k$, $k \ne j_i$, separating $t$, we have that $\langle x_i^k, q\rangle = t \cdot \eta_k$.
Define $M = \max_{k \ne j_i}\big(|\eta_k| \big)$. $M$ is a random variable distributed as the maximum of of $T-1$ i.i.d. Gaussians distributed according to $N(0, 1/d)$.
We bound $M$ by investigating an upper bound of the Gaussian CDF.  Recall that the moment generating function of a Gaussian random variable $X \sim N(0, 1)$ is given by $M_X(r) = \mathbb{E}[e^{rX}] = e^{\frac{1}{2} r^2}$.  Then note that for any $s > 0$  we have 
\[\mathbb{P}(X \geq r) = \mathbb{P}(e^{s X} \geq e^{s r})  \leq e^{-s r} M(s) = e^{-s r + \frac{1}{2} s^2}\]
where the inequality is an application Markov's inequality. Setting $s = r$ this gives the tail bound
\begin{equation}
    \mathbb{P}(X \geq r ) \le \exp(-r^2/2).
\end{equation}
For our settings with $\eta_k \sim N(0, 1/d)$
\begin{equation}
    \PP( \eta_k  \geq r) \leq \exp(-dr^2/2).
\end{equation}
For a two-sided bound, by symmetry of the distribution we have 
\begin{equation}
    \PP( |\eta_k| \geq r) \leq 2\exp(-dr^2/ 2).
\end{equation}
Therefore a union bound results in 
\begin{align*}
    \PP(M \ge r) &= \PP\parens{\bigcup_{k\ne {j_i}} |\eta_k | \ge r} \\
    &\le \sum_{k \ne j_i} \PP( |\eta_k| \ge r)  \\
    &\leq (T-1) \cdot  2\exp(-dr^2/2) \\
    &\leq 2 T \exp(-dr^2/2).
\end{align*}
We can bound the bulk of the distribution of $M$ as
\begin{equation}
    \PP(M  < r) \ge 1 - 2T \exp(-dr^2/2).
\end{equation}
Taking $r = c_{\calD}/4$, then with probability at least $1 -  2T \exp(-d(c_{\calD}/4)^2/2) = 1 - 2T \exp(-d(c_\calD)^2/32)$ we have
\begin{equation}\label{eq:M_bound}
M = \max_{k \neq j_i} |\eta_k| < \frac{c_\calD}{4}
\end{equation}
and thus
\begin{equation}\label{eq:max_token_proj_bound}
    \max_{k \neq j_i}(\langle q, x_i^k \rangle ) < \frac{t c_{\calD}}{4}.
\end{equation}
With high probability, \cref{eq:M_bound} holds.  This implies that for $j_i$,
\begin{align*}
    S_{j_i} &= \frac{\exp(\langle x_i^{j_i}, q \rangle)}{\sum_{j=1}^{T} \exp (\langle x^j_i, q \rangle)} \\
    &= \frac{1}{1 + \sum_{j \neq j_i}^{T} \exp (\langle x^j_i, q \rangle - \langle x_i^{j_i}, q \rangle))} \\
    &\geq \frac{1}{1 + \sum_{j \neq j_i}^{T} \exp (\langle x^j_i, q \rangle - t c_\calD / 2))} \\
    &\geq \frac{1}{1 + \sum_{j \neq j_i}^{T} \exp (-t c_\calD / 4))} \\
    &= \frac{1}{1 + (T - 1) \exp (-t c_\calD / 4))} \\
    &= 1 - \frac{(T - 1) \exp (-t c_\calD / 4))}{1 + (T - 1) \exp (-t c_\calD / 4))}.
\end{align*}
Note that $c_\calD > 0$ and $T$ is fixed.  Nonetheless,  the probability bound is independent of $t$ which may take arbitrarily values, e.g. for any $\epsilon > 0$, take $t = 4/c_{\calD} \log(T/ \epsilon)$, which gives 
\begin{equation}
    S_{j_i} \ge 1 - \epsilon.
\end{equation}
Since $S_k \ge 0$ for each $k$ and $\sum_{k=1}^{T} S_k = 1$ we have for $k \ne j_i$,
\begin{equation}
    S_k \le \sum_{j \ne j_i} S_j = 1 - S_{j_i} \le \epsilon.
\end{equation}
We consider the classifer prediction
\begin{equation}
    f(\xb; q, u) = \sigma \parens{\sum_{j=1}^{T}  S_j \langle u, x_i^j \rangle + b}
\end{equation}
where $b$ is a bias parameter we can choose.  Recall $u=w_\calD$.  Focusing on the inside of the sign function
\begin{align*}
    \sum_{j=1}^{T}  S_j \langle w_\calD, x_i^j \rangle &= S_{j_i} \langle w_\calD, x_i^{j_i} \rangle  + \sum_{k\ne j_i} S_k \eta_k \\
    &\ge S_{j_i} \langle w_\calD, x_i^{j_i} \rangle - \sum_{k \ne j_i} S_k \max_{j \neq j_i}(|\eta_j|) \\
    &\ge (1-\epsilon)c_{\calD} / 2 - \epsilon \cdot c_{\calD}/4 \\
    &= (1- (3/2)\epsilon )c_{\calD} / 2 > \frac{c_\calD}{4}
\end{align*}
provided that $\epsilon < 1/3$.  If we take $b = -\frac{c_{\calD}}{4}$ we have that for $q = t w_\calD$ and $u = w_\calD$
\begin{equation}
    f(\xb_i; q, u, b) = \sigma\parens{  \sum_{j=1}^{T}  S_j \langle w_\calD, x_i^j \rangle  + b} = 1  = y_i.
\end{equation}
Next we address the case where $y_i=-1$.
We consider the classifier prediction
\begin{equation}
    f(\xb; q, u, b) = \sigma \parens{\sum_{j=1}^{T}  S_j \langle u, x_i^j \rangle + b}
\end{equation}
with $u=w_\calD$.  Again for $k \neq j_i$ we have that
\begin{equation}
    \max_{k \neq j_i}(\langle u, x_i^k \rangle ) < c_{\calD}/4.
\end{equation}
On the other hand for $j_i$,~$y_i=-1$ we have that
\begin{gather*}
    y_i(\langle x_i^{j_i}, w_\calD \rangle + b_\calD) \ge c_{\calD} \\ 
    \implies \langle x_i^{j_i}, w_\calD \rangle + b_\calD \le -c_{\calD} \\
    \implies \langle x_i^{j_i}, w_\calD \rangle \le -c_{\calD} / 2 < 0
\end{gather*}
where we have used the hypothesis that $c_\calD \geq 2 |b_\calD|$ in the last line.  We consider the term inside the classifier.  We note that 
\begin{align*}
    \sum_{k=1}^{T}  S_k \langle u, x_i^k \rangle
    &< \sum_{k \neq j_i}^{T}  S_k \langle u, x_i^k \rangle \\
    &< \sum_{k \neq j_i}^{T} S_k \cdot (c_{\calD})/4 \\
    &\leq c_{\calD}/4
\end{align*}
where in the first inequality we have used the fact that $S_{j_i} \langle u, x_i^{j_i} \rangle < 0$ and in the last inequality we have used the fact that $\sum_{j = 1}^T S_j = 1$.  Therefore again with bias term $b = -c_{\calD}/4$ we have that
\begin{equation}
    \sum_{j=1}^{T}  S_j \langle u, x_i^j \rangle   + b  < 0
\end{equation}
and $ f(\xb_i; q, u, b) = -1 = y_i$.  Thus we have just shown that with probability $1 - 2T \exp(-d(c_\calD)^2/32)$ the model $f(x; q, u, b)$ with $u = w_\calD, q = t w_\calD$, $b = - c_\calD / 4$ gives the correct label for $\xb_i$.  Taking the union bound over all $n$ points in $\calD$ we get with probability at least $1 - 2Tn \exp(-d(c_\calD)^2/32) \geq 1 - \delta$ the model $f(x; q, u, b)$ with $u = w_\calD, q = t w_\calD$, $b = - c_\calD / 4$ separates $\calD$.
\par

\paragraph{Negative result for the linear model}
We consider the linear classifier 
\begin{equation}
    g(\xb; w, b) = \sigma\parens{b + \sum_{j=1}^{T} \langle w, x^j \rangle}
\end{equation}
where $w \in \mathbb{R}^d$ is restricted to have unit norm $\|w\| = 1$.  For an input $\xb_i$ under the aggregation, the term inside the sign function simplifies to
\begin{equation*}
    \sum_{j=1}^{T} \langle w, x_i^j \rangle   = \langle w, \sum_{j=1}^{T} x_i^j \rangle.
\end{equation*}
We recall that the $x_i^k$ for $k \ne j_i$ are distributed according to $N(0, \frac{1}{d} I)$.  Thus we have that
\begin{equation}
     \alpha_i := \sum_{k \ne j_i} x_i^k \sim N\parens{0, \frac{T-1}{d}}.
\end{equation}
So the problem of classification is equivalent to learning a linear classifier over the separating tokens under the presence of Gaussian noise with distribution $N(0, \frac{T-1}{d})$.  Let $i^*$ be the index corresponding to the input $\xb_{i^*}$ with smallest margin, i.e.
\[ i^* = \mathrm{argmin}_i y_i(\langle w, x_{i}^{j_i} \rangle + b_\calD). \]
Then we have that $y_{i^*}(\langle w, x_{i^*}^{j_{i^*}} \rangle + b_\calD) \le c_{\mathcal{D}}$.  We note that any linear classifier $g(\xb; w, b)$ with $\norm{w}_2 = 1$ will fail to classify $\calD$ whenever $y_{i^*} \alpha_{i^*} < -c_\calD$.  Thus we will lower bound the probability of $\PP(y_{i^*} \alpha_{i^*} < - c_{\calD})$.  Note for a standard Gaussian random variable $\eta \sim N(0,1)$ as shown in \cite{cookjohn}  we have for $r > 0$
\begin{equation}
    \PP(\eta > r) \ge \frac{1}{\sqrt{2\pi}} \frac{r}{r^2 +1} \exp(-r^2/2).
\end{equation}
Set $s = \frac{\sqrt{d}}{\sqrt{T - 1}} c_{\mathcal{D}}$.   Then by symmetry of the Gaussian distribution the above bound translates into the following bound for $\alpha_{i^*}$
\[ \PP(y_{i^*} \alpha_{i^*} < -c_{\mathcal{D}}) \ge \frac{1}{\sqrt{2\pi}} \frac{s}{s^2 +1} \exp(-s^2/2). \]
It follows that for any $w \in \mathbb{R}^d$ such that $\|w\|_2 =1$ that the linear classifier $g(\xb; w, b)$ incorrectly classifies $\calD$ with probability at least
\[ \frac{1}{\sqrt{2\pi}} \frac{s}{s^2 +1} \exp(-s^2/2). \]
This completes the second part of the proof.
\end{proof}

\section{Further Results}\label{sec:add}
We present additional experiments below. In Subsection \ref{subsec:per_dataset} we present per-dataset results for additional architectures and a discussion about ensembling \name is given in Subsection \ref{subssec:ensemble}.

\subsection{Per-dataset results for different architectures as presented in Table \ref{tab:many_arch}}\label{subsec:per_dataset}
Table \ref{tab:many_arch_per_dataset} provides per-dataset results that are presented in aggregate in Table \ref{tab:many_arch}. Below we present the results for ConvNext-Base and ViT-L/16 (original pre-training) pre-trained models (with the results for ViT-L/16 DeiT and SWIN-L presented in Table \ref{tab:fine_grained_vit} and Table \ref{tab:fine_grained_swin} respectively). 
\begin{table*}[ht]
 \centering
   \resizebox{0.8\textwidth}{!}{

  \begin{tabular}{c | c | a ccc}
\toprule
 \multicolumn{6}{c}{\textbf{Top-1 Test Error for \underline{ConvNext-B}}}\\
 \midrule
Dataset &       \makecell{Full \\ fine-tuning} & \name  &      \makecell{ \name~\\ (last)} & Inter. LP & LP  \\
\midrule
          CUB-200 &  9.3 &         9.3 &          9.3 &         13.0 &         13.0 \\
          DTD & 16.7 &        17.4 &         17.4 &         18.6 &         18.6 \\
  Flood Depth & 16.9 &        16.5 &         20.5 &         19.4 &         19.9 \\
      EuroSAT  &  0.9 &         1.6 &          2.2 &          2.8 &          3.1 \\
     Aircrafts & 10.5 &        17.9 &         23.1 &         54.7 &         54.7 \\
     Herbarium & 17.0 &        22.7 &         26.4 &         37.4 &         39.5 \\
        MIT-67 & 10.9 &        10.3 &         10.7 &         10.2 &         10.4 \\
Oxford Flowers &  0.5 &         0.4 &          0.4 &          0.6 &          0.6 \\
  Oxford Pets &  5.2 &         4.6 &          5.6 &          5.9 &          6.0 \\
  Stanf. Cars &  6.8 &         9.3 &         14.4 &         39.9 &         39.9 \\
  Stanf. Dogs &  8.9 &         7.6 &          7.6 &          7.3 &          7.3 \\
  \midrule
  \makecell{Ave. Top-1 Test Error}   &  9.4 & 10.7 (-7.4) & 12.8 (-12.6) & 19.7 (-44.2) & 20.0 (-44.2) \\
\bottomrule
\end{tabular}
}

   \resizebox{0.8\textwidth}{!}{

  \begin{tabular}{c | c | a ccc}
\toprule
 \multicolumn{6}{c}{\textbf{Top-1 Test Error for \underline{ViT-L/16} (ViT pre-training)}}\\
 \midrule
Dataset &       \makecell{Full \\ fine-tuning} & \name  &      \makecell{ \name~\\ (last)} & Inter. LP & LP  \\
\midrule
       CUB-200 & 11.7 &        10.9 &         10.9 &         12.2 &         12.2 \\
           DTD & 18.3 &        18.9 &         20.1 &         19.9 &         20.1 \\
   Flood Depth & 20.8 &        18.1 &         18.7 &         18.7 &         18.7 \\
      EuroSAT  &  0.8 &         1.1 &          1.9 &          2.5 &          3.5 \\
     Aircrafts & 20.7 &        23.2 &         28.2 &         44.5 &         46.4 \\
     Herbarium & 20.3 &        26.9 &         31.3 &         38.9 &         41.3 \\
        MIT-67 & 12.8 &        11.3 &         11.9 &         10.4 &         11.1 \\
Oxford Flowers &  0.6 &         0.3 &          0.4 &          0.3 &          0.4 \\
   Oxford Pets &  5.5 &         5.3 &          5.4 &          6.5 &          6.5 \\
   Stanf. Cars &  9.3 &        10.9 &         12.9 &         27.6 &         30.2 \\
   Stanf. Dogs & 11.0 &        10.4 &         10.4 &         10.1 &         10.1 \\
         \midrule
   \makecell{Ave. Top-1 Test Error}       & 12.0 & 12.5 (-6.6) & 13.8 (-11.0) & 17.4 (-23.8) & 18.2 (-25.7) \\
\bottomrule
\end{tabular}
}

\vspace{0.5cm}

\centering
\caption{\textbf{Per-dataset Adaptation Top-1 Test Error on various architectures} We test transfer learning performance of fine-grained datasets applied to different architectures and pre-trainings including, ViTs, SWIN, and convolutional networks. We report the per-dataset Top-1 test error for the 11 datasets presented in Table \ref{tab:many_arch}}\label{tab:many_arch_per_dataset}.
\end{table*}

\subsection{Ensembling learned adapters}\label{subssec:ensemble}
\begin{figure}[ht]
    \centering
    \includegraphics[width=8cm]{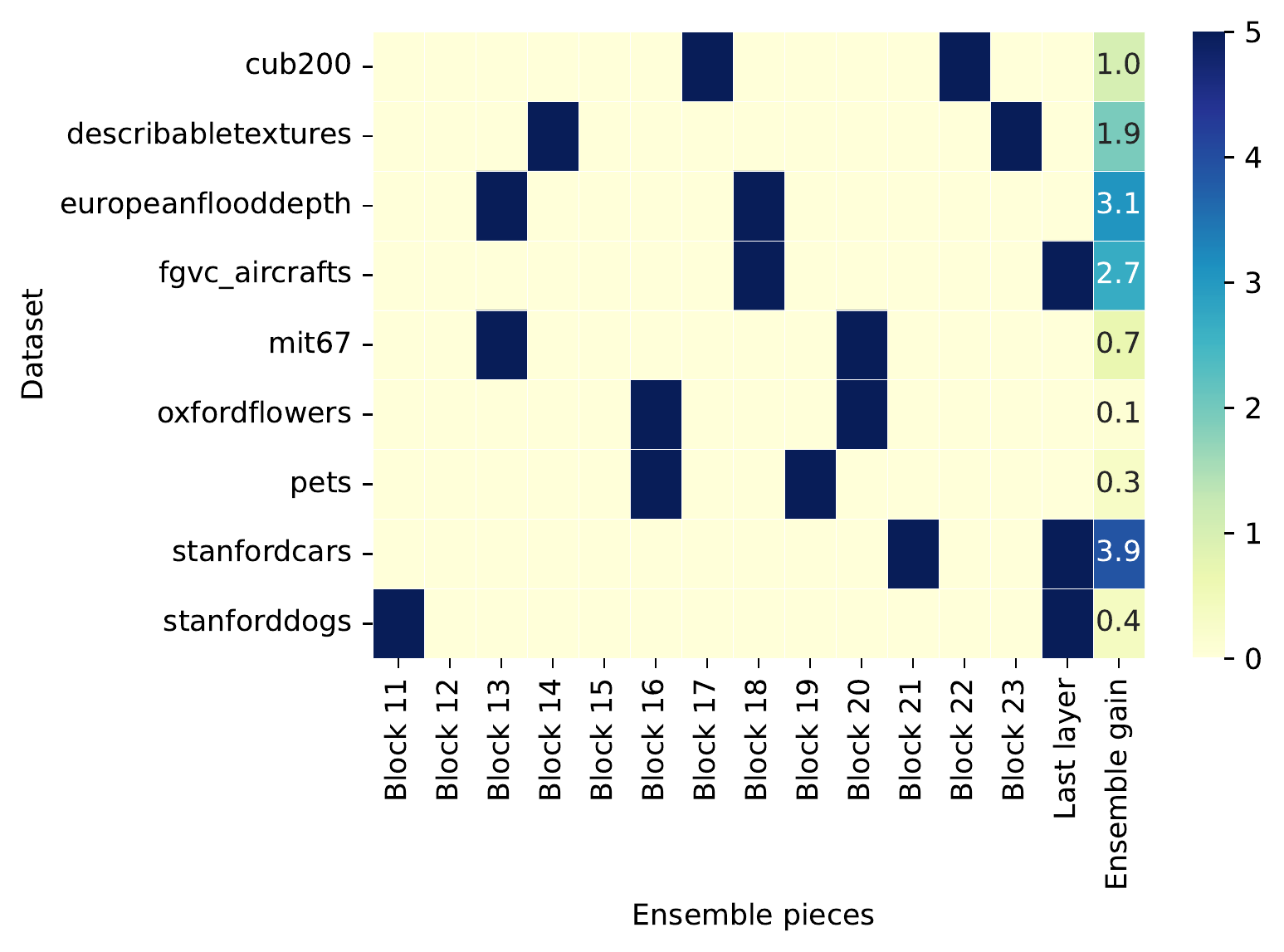}
    \caption{\textbf{Optimal Representation pairings} Optimal ensemble pairs of \name of listeners at different locations of the network; Optimal ensembles can improve over any single layer. ViT-L/16 DeiT pre-training.}
\label{fig:ensemble}
\end{figure}

Because of ``one-to-many'' inference of \name can take a set of independently learned adapters and ensemble them without a marginal increase to the inference cost. We follow non-parametric equal-weight ensembling, 
by taking the output predictions of two adapters $h_1(x), h_2(x)$ on a sample image $x$. Note that the adapters are computed with their relevant representations via a single forward pass, which makes the execution of $h_1(x)$ and $h_2(x)$ together only incrementally higher than computing just $h_1(x)$. The ensemble is defined as
\begin{equation}
    h^{*}(x) = \frac{h_1(x) + h_2(x)}{2}.
\end{equation}
Given the large combinatorial selection of $k$ adapters from the $l$ learned adapters we consider the case of ensembling with just two adapter members. After training we evaluate all $m(m-1)/2$ such pairs and compare them with the top performing single layer predictor which we present in Figure \ref{fig:ensemble}.  In the figure, we illustrate the representations and corresponding adapter pairs that lead to best performance and also present the computed ensemble gain which is the difference between the ensembled model accuracy and the top accuracy of any single adapter.

In addition to improving classification accuracy, ensembling can aid in improving robustness and out of distribution performance which we leave as a future work. Further directions of ensembling include ensembling performance when using adapters of different adapter architectures (e.g. an MLP-3 ensembled with an \name adapter) or adapters that use representations from different neural networks  \cite{gontijo2021no}.

\subsection{Ablation on the number of queries}
We apply an ablation to see the effects of using a different number of queries in the InCA adapter architecture. In particular, the InCA adapter is written as,
\begin{align*}
v_\text{cross}(\mathbf{z})_{[1:m]}  &:= \operatorname{cross-attn}_\theta([z^1, \ldots, z^T], [q_1, \dots q_m])\\
\inca_\theta(\mathbf{z}) &:=
\operatorname{head}_\theta \circ \operatorname{norm} \,  (\operatorname{avg-pool} (v_\text{cross}(\mathbf{z})_{[1:m]}).
\end{align*}
For $m > 1$ the output of tokens $[q_1, \dots q_m]$ through the $\operatorname{cross-attn}$ layer are averaged, and we test whether using $m >1$ brings additional representational benefit to each adapter. We present the result in Table \ref{tab:query_ablation} and observe that using a different $m$ does not have a consistent effect on the accuracy of the learned adapters, and in our experiments we use $m=1$ for InCA adapters to be most computationally efficient.

\begin{table}[ht]
    \centering
    \
   \resizebox{0.5\textwidth}{!}{
\begin{tabular}{c|cccc}
\toprule
                \multicolumn{5}{c}{\textbf{Top-1 Test Error for ViT-L/16 (DeiT pre-training)}}   \\
\midrule
               & \multicolumn{4}{c}{$\#$ of InCA queries ($m$)}   \\
\midrule
       Dataset &                 1 &    2 &    4 &   16 \\
\midrule

       CUB-200 &               9.1 &  9.5 &  9.6 &  9.5 \\
           DTD &              17.8 & 18.4 & 19.2 & 19.1 \\
     Aircrafts &              15.8 & 19.3 & 19.8 & 16.8 \\
        MIT-67 &              10.1 & 10.8 &   11.0 & 10.9 \\
Oxford Flowers &               0.3 &  0.3 &  0.3 &  0.4 \\
   Oxford Pets &               4.7 &  4.7 &  4.5 &  4.4 \\
   Stanf. Cars &               8.4 &  8.7 &  8.8 &  8.2 \\
   Stanf. Dogs &               8.1 &  6.3 &  6.3 &  5.9 \\
\bottomrule
\end{tabular}   
}
\vspace{0.2cm}
\caption{\textbf{Varying $\#$ of queries in the InCA adapter} We run an ablation testing the effect of applying a different number of queries $q_1, \dots q_m$ and then averaging when using the InCA adapter. We observe that in most cases $m$ does not have a big effect on accuracy and that $m=1$ has sufficient representation capacity for the adapter.}
    \label{tab:query_ablation}
\end{table}

 \section{Implementation details}\label{sec:implement}
We present the optimization and augmentation details for training \name, and note we use standardized procedures for augmentation and training (without extensive hyper-parameter optimization) of the different transfer learning methods we evaluate.

\paragraph{Augmentation}
Unless otherwise specified we train with input image size 224 and standard augmentation practice \cite{radford2018improving}. In particular, during training we resize to image-size 256 and apply random cropping, for testing we apply resizing and center cropping. For larger image resolutions we maintain the same resize-crop ratio of $0.875$.

\paragraph{Optimization}
For the linear probing and \name approaches, we train with the AdamW optimizer \cite{loshchilov2017decoupled}, cosine annealing learning rate scheduler \cite{loshchilov2016sgdr} for 30 epochs and with weight decay 1e$-4$. In each method we sweep over 2 learning rates lr = \{1e$-4$,3e$-4$\}.
For full fine-tuning, we also train with AdamW optimizer (weight decay 1e$-4$), cosine annealing for 30 epochs, but in addition, identify optimal learning rates for each pre-training and architecture separately. We first identify an architecture coarse-range learning rate based on performance on 5 datasets by sweeping over lr = \{1e$-2$,1e$-3$,1e$-4$,1e$-5$,1e$-6$\} followed by a refined sweep with learning rates lr = B,~2B with B being the optimal coarse learning rate.

For the VPT baseline, we follow the details presented in the paper and train with VPT-Deep which was observed to outperforms VPT-Shallow. To train VPT, we use the SGD optimizer with momentum and cosine annealing for 100 epochs. For each dataset we run a sweep on the prompt length \{5,20,100\}, base learning rate \{0.25,0.1,0.05,0.01\}, and weight-decay \{1e$-2$,1e$-4$\} for a total of 24 runs with 100-epochs for each dataset. We compare the training cost of \name and VPT-Deep in Table \ref{tab:inca_vpt_eff}. In general we note that the shallow and small architecture of \name or linear probing that are separate from the base model makes them straightforward to optimize, compared with adaptation methods that receive back-propagated gradients from a frozen intermediate layer of the network as shown in \cref{fig:cal_diagram}.

For the LoRA baseline \cite{hu2021lora} we apply a LoRA modified attention to each block's self-attention layer ($W_k,W_q,W_v$) in ViT based architectures and to each block's WindowAttention for SWIN. For the low rank dimension we sweep over the best value among $d=5,10,50$. For BitFit we follow the discussion in \cite{zaken2021bitfit} and train all of the bias-parameters in the network in addition to full training of the head. Analogously for \cite{li2016revisiting} we follow their procedure with LayerNorm which includes training each of the LayerNorm parameters ($\gamma, \beta$) for each layer along with training of the head of the pre-trained model. For all of the efficient training methods above we sweep over lr=\{3e$-5$,1e$-4$,3e$-4$,1e$-3$\} to identify the best learning rate for the dataset.

\paragraph{Broader Impacts}
Our method, InCA enables efficient and modular model adaptation that can be applied to any strong available pre-trained backbone. In that sense, InCA reduces the computational barriers to entry for training and evaluating over a large set of (potentially massive scale) models and optimization settings to identify a model to be used for downstream adaptation. This bridges the gap between cutting edge research in general visual representation learning and specific domain applications, especially since the best performing models are computationally expensive to adapt.
Given that InCA operates well on fine-grained visual datasets, this can have positive applications in scientific domains such as medical imaging. In many scientific domains, the available datasets are known to be fine-grained yet also with sparse training data. In addition the ease of use and reduced computational costs associated with downstream adaptation with InCA makes it possible for domain experts without machine learning expertise to use InCA without access to large computational resources. This can enable domain researchers solve their domain problems by leveraging various public pre-trained models to achieve competitive results.

\end{document}